\documentclass[reqno,11pt]{article} 
\usepackage[utf8]{inputenc}

\title{\textbf{Privacy Preserving Off-Policy Evaluation}}

\author{Tengyang Xie \qquad Philip S. Thomas \qquad Gerome Miklau\\
College of Information and Computer Sciences, UMass Amherst\\
\texttt{\{txie,pthomas,miklau\}@cs.umass.edu}}

\date{}

\usepackage{graphicx}
\usepackage{geometry}

\usepackage{amsmath}
\numberwithin{equation}{section} 

\usepackage{amssymb}
\usepackage{amsfonts}
\usepackage[table,xcdraw]{xcolor}
\definecolor{darkblue}{HTML}{000080}

\usepackage{url}
\usepackage[colorlinks=true, allcolors=blue]{hyperref}

\usepackage{mathtools}

\usepackage{lineno}

\usepackage{mathrsfs}
\mathtoolsset{showonlyrefs}

\usepackage{natbib}
\setcitestyle{authoryear} 
\setcitestyle{square}
\setcitestyle{semicolon} 

\usepackage[labelformat=simple]{subcaption}

\usepackage{algorithm}
\usepackage{algorithmic}

\usepackage{amsthm}
\newtheorem{theorem}{Theorem}
\newtheorem{lemma}{Lemma}

\newtheorem{definition}{Definition}
\newtheorem{assumption}{Assumption}
\newtheorem{remark}{Remark}

\usepackage[inline,ignoremode]{trackchanges}

\addeditor{tx}

\newcounter{RomanNumber}
\newcommand{\RomanNum}[1]{\setcounter{RomanNumber}{#1}\Roman{RomanNumber}}

\begin{document}
	
\maketitle

\begin{abstract}
Many reinforcement learning applications involve the use of data that is sensitive, such as medical records of patients or financial information.
However, most current reinforcement learning methods can leak information contained within the (possibly sensitive) data on which they are trained.
%
%
To address this problem, we present the first differentially private approach for off-policy evaluation. 
We provide a theoretical analysis of the privacy-preserving properties of our algorithm and analyze its utility (speed of convergence).
After describing some results of this theoretical analysis, we show empirically that our method outperforms previous methods (which are restricted to the on-policy setting). 
\end{abstract}


\section{Introduction}
Many proposed applications of \textit{reinforcement learning} (RL) involve the use of data that could contain sensitive information. 
For example, \citet{raghu2017deep} proposed an application of RL and off-policy evaluation methods that uses peoples' medical records, and \citet{theocharous2015personalized} applied off-policy evaluation methods to user data collected by a bank in order to improve the targeting of advertisements. 
In examples like these, the data used by the RL systems is sensitive, and one should ensure that the methods applied to the data do not leak any sensitive information. 


Recently, \citet{balle2016differentially} showed how techniques from \textit{differential privacy} can be used to ensure that (with high probability) policy evaluation methods for RL do not leak (much) sensitive information. 
In this paper we extend their work in two ways. 
First, RL methods are often applied to batches of data collected from the use of a currently deployed policy. 
The goal of these RL methods is not to evaluate the performance of the current policy, but to improve upon it. 
Thus, policy evaluation methods must be \textit{off-policy}---they must use the data from the behavior policy to reason about the performance of newly proposed policies. 
This is the problem of \textit{off-policy evaluation}, and both of the previous medical and banking examples require these methods. 
%
%
Whereas \citet{balle2016differentially} consider the \textit{on-policy} setting (evaluating the deployed policy), we focus on the off-policy setting.

Second, \citet{balle2016differentially} achieve their privacy guarantee using output perturbation:
they first run an existing (non-private) least-squares policy evaluation method, resulting in a real-valued vector; then they add random noise to each element of the vector. 
Although this approach was one of the first and most simple methods for ensuring that guarantees of privacy hold \citep{dwork2006calibrating}, more sophisticated methods for ensuring privacy have since been developed. 
We show how one of these newer approaches to differential privacy, which adds noise to stochastic gradient descent updates \citep{song2013stochastic, bassily2014differentially}, rather than to the least squares solution, can be combined with GTD2, the dominant off-policy evaluation algorithm \citep{sutton2009fast}.

After presenting our new privacy preserving off-policy evaluation algorithm, which we call \textit{gradient perturbed off-policy evaluation} (GPOPE) to differentiate it from the previous output-perturbation methods, we provide proofs of privacy and convergence rate. 
We use the properties of R\'enyi differential privacy and its amplification via subsampling \citep{bun2016concentrated,mironov2017renyi,balle2018privacy,wang2018subsampled} together with the moments accountant technique \citep{abadi2016deep} to effectively keep track of $(\varepsilon,\delta)$-differential privacy parameters through all steps of our algorithm.
The convergence rate analysis quantifies the trade-off between the strength of the privacy guarantees that our algorithms provide and the accuracy of their off-policy predictions. 


Since the on-policy setting is a special case of the off-policy setting (where the policy being evaluated happens to be the same as the currently deployed policy), we  can compare our algorithm directly to the output-perturbation methods of \citet{balle2016differentially} in the on-policy setting. 
We show empirically that our algorithm offers greater utility, i.e., using the same data, our algorithm can provide stronger guarantees of privacy for the same degree of prediction error.
%
We also conduct experiments in the off-policy setting, where prior work is not applicable, and the results support the conclusions of our analytic analysis. 

The rest of the paper is organized as follows. 
We review the relevant background on off-policy evaluation in Section \ref{sec:bgope} and background on differential privacy in Section \ref{sec:bgpriv}. 
We present our algorithm in Section \ref{sec:algo}. 
In Section \ref{sec:privana} we analyze the privacy preserving properties of our algorithm, and in Section \ref{sec:utiana} we provide an analysis of the utility of our algorithm. 
We provide an empirical case study in Section \ref{sec:exp}, using a synthetic MDP that mimics characteristics of a medical application, the standard Mountain Car domain, and a more challenging HIV simulator. We conclude in Section \ref{sec:discussion} with a discussion of future work.


\section{Background: Off-Policy Evaluation}
\label{sec:bgope}


This section offers a brief overview of off-policy evaluation, including the definition of Markov decision processes, mean squared projected Bellman error, and the saddle-point formulation of the gradient temporal-difference (GTD2) off-policy evaluation method \citep{sutton2009fast}. 




A \emph{Markov decision process} (MDP) \citep{sutton1998reinforcement,puterman2014markov} is a tuple $(\mathcal{S}$, $\mathcal{A}$, $\mathcal P$, $\mathcal R$, $\gamma)$, where $\mathcal{S}$ is the finite set of possible states, $t \in \{0,1,2,\dotsc\}$ is the \textit{time step}, $S_t$ is the state at time $t$ (a random variable), $\mathcal{A}$ is the finite set of possible actions, $A_t$ is the action at time $t$, $\mathcal P:\mathcal S \times \mathcal A \times \mathcal S \to [0,1]$ is the \textit{transition function}, defined such that $\mathcal  P(s,a,s')\coloneqq \Pr(S_{t+1}=s'|S_t=s,A_t=a)$, $R_t$ is the scalar reward at time $t$, $\mathcal{R}: \mathcal{S} \times \mathcal{A} \rightarrow \mathbb{R}$ is defined such that $R(s,a)\coloneqq \mathbf{E}[R_t|S_t=s,A_t=a]$, and $\gamma \in [0,1]$ is a parameter that characterizes how rewards are discounted over time. A policy, $\pi: \mathcal{S} \times \mathcal{A} \to [0,1]$, describes one way that actions can be chosen: $\pi(s,a)\coloneqq\Pr(A_t=a|S_t=s)$.

A key step in many RL algorithms is to estimate the state-value function $V^{\pi}: \mathcal{S} \rightarrow \mathbb{R}$ of a given policy $\pi$, which is defined as $V^{\pi}(s) \coloneqq \mathbf{E}[\sum_{t = 0}^{\infty}\gamma^{t}R_t|S_0 = s,\pi]$. 
The process of estimating a state-value function is known as \textit{policy evaluation}. 
In this paper we consider the problem of \textit{off-policy} evaluation, wherein we estimate $V^\pi$ given data (states, actions, and rewards) sampled from applying a different policy, $\pi_b$, called the \textit{behavior policy}, which may be different from $\pi$ (i.e., the policy being evaluated). 
Furthermore, we consider the setting where a linear function approximator, $\widehat V^\pi$, is used. 
That is $\widehat V^\pi$ can be written as $\widehat V^\pi(s)\coloneqq \theta^\intercal \phi(s)$, where $\theta \in \mathbb R ^n$ is a set of weights, and $\phi(s)\in \mathbb R ^n$ is a feature vector associated with state $s$. 

Let $H \coloneqq \{(S_t,A_t,R_t,S_{t + 1})\}_{t = 0}^{\tau}$ be a \textit{trajectory} with length $\tau$. Often each trajectory contains data pertaining to a single individual over time. In real-world applications, states often describe people: their bank balance when the MDP models automatic selection of online credit card ads \citep{theocharous2015personalized}, or medical conditions when selecting between treatments \citep{raghu2017deep}. Similarly, actions can include drug prescriptions and rewards can encode medical outcomes. 

Recent work has shown that optimizing the weight vector, $\theta$, can be phrased as a saddle point problem \citep{liu2015finite}: 
$
\min_\theta\max_w  \mathcal{L}(\theta, w),
$
where
\begin{equation}
\begin{aligned}
& \mathcal{L}(\theta, w) \coloneqq  w^\intercal (b - A\theta) - \dfrac{1}{2}\|w\|_C^2\\
\label{eq:abc}
& A \coloneqq \mathbf{E}\left [\frac{1}{\tau}\sum_{t=0}^\tau \rho_t \phi_t (\phi_t - \gamma\phi_{t+1})^\intercal \right ], \\
& b \coloneqq \mathbf{E}\left [\frac{1}{\tau}\sum_{t=0}^\tau \rho_t \phi_t R_t\right ], ~C \coloneqq \mathbf{E}\left [\frac{1}{\tau}\sum_{t=0}^\tau \phi_t \phi_t^\intercal\right ],
\end{aligned}
\end{equation}
where
$w \in \mathbb R^n$ is introduced by duality \citep{boyd2011distributed,sutton2009fast,liu2015finite},
the expected values in \eqref{eq:abc}, are over states, $S_t$, actions, $A_t$, and rewards, $R_t$, produced by running the behavior policy, $\pi_b$, $\tau$ is the (finite) length of the trajectory, $\phi_t$ is shorthand for $\phi(S_t)$, and $\rho_t \coloneqq  \pi(S_t,A_t) / \pi_b(S_t,A_t)$.

\citet{liu2015finite} proposed using a stochastic gradient method to optimize this saddle-point problem. 
This algorithm uses the following unbiased estimates of $A, b,$ and $C$, produced using the states, actions, and rewards from the $i^\text{th}$ trajectory (which is of length $\tau_i$):
\begin{equation}
\begin{aligned}
\label{eq:atbtct}
& \widehat{A}_{i} = \dfrac{1}{\tau_{i}}\sum_{t = 0}^{\tau_{i}}\rho_t \phi_t(\phi_t-\gamma \phi_{t+1})^\intercal, \\
& \widehat{b}_{i} =  \dfrac{1}{\tau_{i}}\sum_{t = 0}^{\tau_{i}}\rho_t \phi_t R_t, ~ \widehat{C}_{i} = \dfrac{1}{\tau_{i}}\sum_{t = 0}^{\tau_{i}}\phi_t\phi_t^\intercal.
\end{aligned}
\end{equation}
The resulting stochastic gradient algorithm proposed by \citet{liu2015finite} is identical to the GTD2 algorithm, and is given by the following update equations:\footnote{Although \citet{sutton2009fast} were the first to derive GTD2, they did not derive it as a stochastic gradient algorithm. \citet{liu2015finite} were the first to show that GTD2 can be phrased as presented here---as a stochastic gradient algorithm for a saddle-point problem.}
\begin{equation}
\begin{aligned}
\label{eq:bos}
& \theta_{i + 1} =  \theta_i + \beta_i \widehat A_i^\intercal w_i,\\
& w_{i + 1} =  w_{i} + \beta_i (\widehat b_i - \widehat A_i\theta_i - \widehat C_i w_i),
\end{aligned}
\end{equation}
where $\beta_1,\beta_2,\dotsc$ is a sequence of positive step sizes.

\section{Background: Differential Privacy}
\label{sec:bgpriv}


In this section we define \textit{differential privacy} (DP) and its application to the data underlying off-policy evaluation. We also describe some tools that aid in analyzing the privacy loss when using gradient methods. 


A \textit{data set}, $d$, consists of a set of $m$ points, $\{x_1,\dotsc,x_m\}$, where each point 
is an element of universe $\mathcal D$ (for RL, a point will correspond to a trajectory, $H$, containing data associated with one person).  For RL applications to human data, each point typically describes a trajectory consisting of a finite sequence of transitions of a single individual, i.e., $x_i = \{(S_t,A_t,R_t,S_{t + 1})\}_{t = 0}^{\tau_{i}}$, and the length of trajectory may vary across individuals.
We assume each trajectory is generated by running a behavior policy, $\pi_b$, and that states, actions, and rewards may all be potentially sensitive and therefore worthy of privacy protection.  We denote by $\mathcal D$ the set of all possible data sets. 

The privacy condition our algorithm provides constrains the treatment of pairs of {\em adjacent} datasets:
\begin{definition}[\textbf{Adjacent Data Set}]
Two data sets, $d, d' \in \mathcal D$ are adjacent if they differ by exactly one point.
\end{definition}

Differential privacy is a formal notion of privacy, which guarantees that the output of a computation on a sensitive data set cannot reveal too much about any one individual.  
Formally, consider a \textit{randomized mechanism}, $\mathcal M$, which takes as input a data set and produces as output an element of some set, $\mathcal Y$. 
%

%
\begin{definition}[\textbf{Differential Privacy}]
\label{def:dp}
Let $\mathcal{M}$ denote a randomized mechanism that has domain $\mathcal{D}$ and range $\mathcal{Y}$. $\mathcal{M}$ satisfies $(\varepsilon, \delta)$-differential privacy for some $\varepsilon, \delta > 0$ , if for every pair of adjacent data sets, $d, d' \in \mathcal{D}$ , and for every $S \subseteq \mathcal{Y}$ the following holds:
\begin{equation}
\Pr\left (\mathcal{M}(d) \in S\right ) \leq e^{\varepsilon}\Pr \left (\mathcal{M}(d') \in S\right ) + \delta.
\end{equation}
\end{definition}

This definition requires that the difference in output probabilities resulting from changing the database by altering any one individual's contribution will be small.  Note that adjacent databases differ in an individual's {\em full} trajectory, not merely one transition.

Applied to our reinforcement learning problem, a differentially private training mechanism allows the public release of a parameter vector of the value function with a strong guarantee: by analyzing the output, an adversary is severely limited in what they can learn about any individual, even if they have access to arbitrary public information.  

Standard $\varepsilon$-differential privacy \citep{dwork2006calibrating} corresponds to $\delta=0$; we use the common relaxation, $(\varepsilon,\delta)$-differential privacy~\citep{dwork2006our,dwork2014algorithmic}.

\section{Differentially Private Off-Policy Evaluation Algorithms}
\label{sec:algo}

In this section we provide the details of our differentially private off-policy evaluation algorithms.

We construct our differentially private off-policy evaluation algorithm by using the Gaussian mechanism \citep{dwork2006calibrating} and the moments accountant \citep{abadi2016deep} to privatize the stochastic gradient off-policy evaluation algorithm presented in \eqref{eq:bos}. 
This involves three steps. 
First, a trajectory of data is collected from running the behavior policy, $\pi_b$. 
Second, a primal-dual stochastic gradient estimate is generated from this data, and its $l_2$ norm is clipped to ensure that it is bounded below a positive constant, $h$. 
Third, we add normally distributed noise to each term of the gradient before updating the weights using the (clipped and noisy) stochastic gradient estimate. 
In subsequent sections we show that the amount of noise that we introduce provides the desired privacy preserving guarantees, regardless of the value chosen for $h$.

Before providing pseudocode for our algorithms, we first define the primal-dual gradient at the $i$-th step, $B_i(\theta,w)$, which is obtained by stacking the estimated primal and negative dual gradients:
\begin{equation}
\begin{aligned}
\label{eq:pdgrad}
B_i(\theta,w) & \coloneqq   
 \left[
 \begin{matrix}
 \dfrac{\partial \widehat {\mathcal L}_i(\theta,w)}{\partial \theta}^\intercal, - \dfrac{\partial \widehat {\mathcal L}_i(\theta,w)}{\partial w}^\intercal
  \end{matrix}
  \right]^\intercal \\
& = \left[
\begin{matrix}
   0 & -\widehat{A}_i^\intercal \\
   \widehat{A}_i & \widehat{C}_i
  \end{matrix}
\right]
\left[
 \begin{matrix}
   \theta \\
   w
  \end{matrix}
  \right] - 
 \left[
 \begin{matrix}
   0 \\
   \widehat{b}_i
  \end{matrix}
  \right],
\end{aligned}
\end{equation}
where
$
\widehat {\mathcal L}_i(\theta, w) \coloneqq w^\intercal (\widehat{b}_i - \widehat{A}_i\theta) - 0.5\|w\|_{\widehat{C}_i}^2,
$
and $\widehat{A}_i,\widehat{b}_i,\widehat{C}_i$ are defined in \eqref{eq:atbtct}. Let $B(\theta,w)$ denote the true primal-dual gradient, $B(\theta,w) \coloneqq \mathbf E [B_i(\theta,w)]$, where the expected values are over states, actions, and rewards produced by running the behavior policy. 





\begin{algorithm*}[t]
\caption{Gradient Perturbed Off-Policy Evaluation (GPOPE)}
\label{alg:DPEPDG}
{\bfseries Input:} Initial point, $(\theta_1,w_1)$, step size sequence $\{\beta_i\}$, clipping bound, $h \in \mathbb R$, private dataset, $d$, with $m$ trajectories, number of iterations, $N$, and a noise scale $\sigma \in \mathbb R$.
\begin{algorithmic}[1]
  \FOR{$i = 1$ {\bfseries to} $N$}
  \STATE Randomly choose a trajectory from private dataset, $d$.
  \STATE Compute $\widehat{A}_i$, $\widehat{b}_i$, $\widehat{C}_i$ using all transitions from sampled trajectory by \eqref{eq:atbtct}.
  \STATE Compute $B_i(\theta_i,w_i)$ as in \eqref{eq:pdgrad}.
  \STATE Clip $\bar{B}_i(\theta_i,w_i) = {B_i(\theta_i,w_i)}/{\max(1,\|B_i(\theta_i,w_i)\|_2 / h)}$.
\STATE Sample vector $\zeta \sim \mathcal{N}(0,\sigma^2 \mathbf I)$, of length $2n$, and compute $\tilde{B}_i(\theta_i,w_i) = \bar{B}_i(\theta_i,w_i) + h \zeta$.
  \STATE $[\theta_{i+1}^\intercal, w_{i+1}^\intercal]^\intercal = [\theta_{i}^\intercal, w_{i}^\intercal]^\intercal - \beta_i \tilde{B}_i(\theta_i,w_i)$.
  \ENDFOR
\end{algorithmic}
\end{algorithm*}

Pseudocode for our new privacy preserving off-policy evaluation algorithm, which we call \textit{gradient Perturbed off-policy evaluation} (GPOPE), is provided in Algorithm \ref{alg:DPEPDG}.


Notice that in GPOPE we use all of the transitions from trajectory $i$ to create the unbiased estimates of $A$, $b$, and $C$. Alternate algorithms could use data from a single trajectory to create multiple estimates of $A$, $b$, and $C$, and thus could perform multiple gradient updates given one trajectory. However, in preliminary experiments we found that the episodic approach taken by GPOPE (where we use all of the data from the trajectory for one update) performed the best. This is supported by our theoretical analysis, which shows that the trade-offs between number of updates, the variance of updates, and the amount of noise that must be added to updates, favors this episodic approach. 



We use $\sigma^2$ to denote the variance of the Gaussian noise in our algorithm. The choice of $\sigma$ depends on the desired privacy level of the algorithm, as discussed in the next section.




\section{Privacy Analysis}
\label{sec:privana}

In this section we provide a formal privacy analysis for our algorithm.  We adapt the moments accounting introduced by \citet{abadi2016deep} and the recent privacy amplification properties of subsampling mechanisms \citep{bun2016concentrated, balle2018privacy, wang2018subsampled} to bound the privacy loss of a sequence of adaptive mechanisms, and we show that our algorithm is $(\varepsilon,\delta)$-differentially private.

%



\begin{theorem}
\label{thmdp}
Given a data set consisting of $m$ points and fixing the number of iterations, $N$, there exist constants $c_1$ and $c_2$, such that for any $\varepsilon < c_1 N / m^2$, Algorithm \ref{alg:DPEPDG} is $(\varepsilon,\delta)$-differentially private for $ \delta > 0$ if
\begin{align}
\label{eq:setsgm}
\sigma \geq \frac{c_2\sqrt{N \log(1/\delta)}}{m\varepsilon}.
\end{align}
\end{theorem}


The detailed proof of Theorem \ref{thmdp} is in the appendix. In the remainder of this section we provide an outline of the proof of Theorem \ref{thmdp}, which proceeds as follows. We first define the \textit{privacy loss} and the \textit{privacy loss random variable}. We use privacy loss to measure the difference in the probability distribution resulting from running $\mathcal{M}$ on $d$ and $d'$. Bounds on the tails of the privacy loss random variable then imply the privacy condition.

%
%
%

\begin{definition}[\textbf{Privacy Loss}]
Let $\mathcal M$ be a randomized mechanism with domain $\mathcal{D}$ and range $\mathcal{Y}$, and $\mathsf{aux}$ be auxiliary input, $d,d' \in \mathcal{D}$ be a pair of adjacent data sets. For an outcome $o \in \mathcal{Y}$, the privacy loss at $o$ is:
\begin{align}
l(o; \mathcal{M},\mathsf{aux}, d, d') \coloneqq \log\left(\frac{\Pr[\mathcal{M}(\mathsf{aux},d) = o]}{\Pr[\mathcal{M}(\mathsf{aux},d') = o]}\right).
\end{align}
\end{definition}

The auxiliary information, $\mathsf{aux}$, could be any additional information available to the adversary. We use $\mathsf{aux}$ here to model the composition of adaptive mechanisms, where we have a sequence of mechanisms and the $i$-th mechanism, $\mathcal M_i$, could use the output of previous mechanisms, $\mathcal M_1, \dots, \mathcal M_{i - 1}$, as its input.

We define the privacy loss random variable using the outcome sampled from $\mathcal{M}(d)$, as $L(\mathcal{M}, \mathsf{aux}, d, d') = l(\mathcal{M}(d); \mathcal{M}, \mathsf{aux}, d, d')$. 


In order to more precisely analyze the privacy cost of sequences of mechanisms, we use a recent advance in privacy cost accounting called the moments accountant, introduced by \citet{abadi2016deep} and which builds on prior work \citep{bun2016concentrated,dwork2016concentrated}.

\begin{definition}[\textbf{Moments Accountant}]
\label{def:ma}
Let $\mathcal{M}: \mathcal{D} \rightarrow \mathcal{Y}$ be a randomized mechanism and $d$, $d'$ a pair of adjacent databases. The $\lambda^{th}$ moment of the privacy loss random variable $L(\mathcal{M}, \mathsf{aux}, d, d')$ is:
\begin{align}
\alpha_{\mathcal{M}}(\lambda; \mathsf{aux},d, d') \coloneqq\log\mathbf{E}[\exp(\lambda L(\mathcal{M},\mathsf{aux}, d, d'))].
\end{align}

The moments accountant is defined as
\begin{align}
\alpha_{\mathcal{M}}(\lambda) \coloneqq \max_{\mathsf{aux},d, d'}\alpha_{\mathcal{M}}(\lambda;\mathsf{aux}, d, d'),
\end{align}
which bounds the $\lambda$-moment for all possible inputs (i.e., all possible $d,d',\mathsf{aux}$).
\end{definition}

In the following lemma we provide an upper bound on the moments accountant for each iteration in our algorithm. This upper bound on the moments accountant is the key for proving Theorem \ref{thmdp}.

\begin{lemma}
\label{lem:privlem}
Let sensitive dataset d contain $m$ trajectories, and $x_i$ be the sampled trajectory in the $i^{th}$ iteration.
Then the randomized mechanism $\mathcal{M}(d) = \bar{B}_t(x_i) + h\cdot\mathcal{N}(0,\sigma^2 \mathbf{I})$ satisfies
\begin{align}
\alpha_{\mathcal{M}}(\lambda) \leq \frac{\lambda(\lambda + 1)}{2 m^2} \min \left\{ 4\left(e^{1 / \sigma^2} - 1\right), 2 e^{1 / \sigma^2}  \right\},
\end{align}
where $\bar{B}_t(x_i)$ denotes $\bar{B}_t(\theta_i,w_i)$ defined in step 5 of Algorithm \ref{alg:DPEPDG}, and $\mathcal{M}$ returns the noised gradient.
\end{lemma}

We prove Lemma 1 using use the amplification properties for R\'enyi differential privacy via subsampling \citep{bun2016concentrated, mironov2017renyi, wang2018subsampled}. 
We provide a detailed proof in the appendix. The results in Lemma \ref{lem:privlem} are similar to a result of \citet{abadi2016deep} when $\sigma^2$ is large (if $\sigma^2 \geq 1/\ln 2$, $\alpha_{\mathcal{M}}(\lambda) \leq {4 \lambda(\lambda + 1)} / {m^2 \sigma^2}$), but our Lemma \ref{lem:privlem} also covers the regime of small $\sigma$ \citet{abadi2016deep} does not cover. Also note that our definition of adjacent data sets is different from that of \citet{abadi2016deep}. Our approach avoids the need to specify a discrete list of moments ahead of time as required in the moments accountant method of \citet{abadi2016deep}.


Note that our algorithm can guarantee $(\varepsilon,\delta)$-differential privacy when each update only uses data from one trajectory. This is because the length of trajectories are not always the same, and so using data from multiple trajectories would cause Lemma \ref{lem:privlem} to not hold.  However, our privacy analysis holds with the same privacy guarantee for the case when a subset of the transitions of the sampled trajectory are used.  Intuitively, the best choice is to use all transitions of the sampled trajectory; we will justify this in the next section.

\section{Utility Analysis}
\label{sec:utiana}

In this section we present the convergence analysis (utility analysis) of our algorithm. For this analysis, we assume that $h$ is selected to be sufficiently large so that the $l_2$ norm of the gradient estimate is not clipped, i.e., the gradient estimates are sufficiently small (empirically, we found this assumption held across all of our experiments). Also, without loss of generality, we can avoid gradient clipping by scaling the objective function \citep{wang2017differentially}, i.e., changing the basis used for approximation. 

Let $(\varepsilon,\delta)$ be the privacy parameters, $N$ be the total number of iterations of the loop in Algorithm \ref{alg:DPEPDG}, and $m$ be the number of trajectories in the data set. 
The noise added to the gradient of $\theta$ is $\mathcal{N}(0,\sigma^2 \mathbf I)$, where $\mathbf I$ is the $2n \times 2n$ identity matrix and $\sigma$ is the noise scale chosen according to Theorem \ref{thmdp}. 
Let $c$ be a constant defined as $c \coloneqq m^2 \|h^2 \sigma^2 \mathbf{I}\|_F^2 / N$, where $\|\cdot\|_F$ is the Frobenius norm. 
Note that we choose $\sigma$ according to Theorem \ref{thmdp}, i.e., $\sigma \geq {c_2\sqrt{N \log(1/\delta)}} / {m\varepsilon}$, so that we have
\begin{align}
\label{eq:def-c}
c \geq 2 n c_2^2 \log(1 / \delta) / \varepsilon^2,
\end{align}
which does not depend on $m$ and $N$.

First, let the optimal solution be expressed as 
\begin{equation}
\begin{aligned}
\label{eq:optimal}
& \theta^* \coloneqq ({A}^\intercal {C}^{-1}{A})^{-1}{A}^\intercal{C}^{-1}{b}, \\
& w^* \coloneqq {C}^{-1}({b} - {A}^\intercal\theta^*).
\end{aligned}
\end{equation}

In order to analyze the convergence of the algorithm, we examine the difference between the current parameters and the optimal solution. We define a residual vector $\xi_j$, at each iteration $j$, and a useful parameter $Q$, as:
\begin{align}
\label{eq:resvec_Q}
\xi_i \coloneqq
  \left[
 \begin{matrix}
   \theta_{i} - \theta^* \\
   w_{i} - w^*
  \end{matrix}
  \right], \,\,\,\,\,\,
Q = \left[
 \begin{matrix}
   0        & -{A}^\intercal\\
   {A}  & {C}
  \end{matrix}
  \right].
\end{align}

Note that the optimal solution can be expressed as \eqref{eq:optimal}. The first order optimally condition is obtained by setting the gradient to zero, which is satisfied by $(\theta^*,w^*)$, such that
\begin{align}
 \left[
 \begin{matrix}
   0        & -{A}^\intercal\\
   {A}  & {C}
  \end{matrix}
  \right]
 \left[
 \begin{matrix}
   \theta^* \\
   w^*
  \end{matrix}
  \right]
=  \left[
 \begin{matrix}
   0 \\
   {b}
  \end{matrix}
  \right].
\end{align}

We have defined $B_i(\theta_i,w_i)$ to be the stochastic approximate gradient at iteration $i$, which is stacking of the approximate primal and negative dual gradient using the $\widehat{A_i},\widehat{b_i},\widehat{C_i}$ at iteration $i$, and 
$B(\theta_i,w_i)$ using the true gradient at iteration $i$. Also let $\tilde B_i(\theta_i,w_i)$ be the perturbed approximate gradient, which is defined in step 7 of Algorithm \ref{alg:DPEPDG}. 

We also define $\Delta_i$ to be the approximation error of the primal-dual gradient at iteration $i$, which is
$
\Delta_i \coloneqq \tilde B_i(\theta_i,w_i) - B(\theta_i,w_i). 
$
Note that $\mathbf{E}[\Delta_i] = 0$, since it is an unbiased stochastic approximation. We introduce an assumption, which ensures that the variance of $\Delta_i$ is bounded:
\begin{assumption}
\label{asm3}
There exists a constant, $G^2$, such that for any $t$,
\begin{align}
\mathbf{E}[\|\Delta_i\|_2^2] \leq G^2 + cN / m^2.
\end{align}
\end{assumption}

\begin{remark}
Note that bounded variance of the stochastic approximation is a standard assumption in the literature of stochastic gradient methods. In our differentially private case, the variance bound should be in terms of the privacy guarantee (i.e. $\varepsilon$ and $\delta$), since Algorithm \ref{alg:DPEPDG} adds normally distributed noise to each term of the gradient. The term $cN / m^2$ in the assumption above follows $cN / m^2 = \|h^2 \sigma^2 \mathbf{I}\|_F^2$, where $\|h^2 \sigma^2 \mathbf{I}\|_F^2$ is the Frobenius norm of the covariance matrix of added noise. In the non-private case, $c$ should be $0$, i.e., $\mathbf{E}[\| B_i(\theta_i,w_i) - B(\theta_i,w_i)\|_2^2] \leq G^2$. 
\end{remark}

Thus, we obtain the key properties of each iteration in our algorithm.

\begin{lemma}
\label{lem:iteana}
Let $\xi_{i + 1}$ be generated by the non-private algorithm at iteration $i$, if we define $Q$ as \eqref{eq:resvec_Q}, and we use $\lambda_{\min}(Q)$ to denote the minimum eigenvalue of $Q$, $\lambda_{\max}(Q)$ to denote the maximum eigenvalue of $Q$. If we choose $\beta_i \leq 1 / \lambda_{\max}(Q)$, we then have
\begin{equation}
\begin{aligned}
\label{eq:finalineq}
& \mathbf{E}[\|\xi_{t + 1}\|_2^2]  \\
\leq & \left(1 - \beta_i\lambda_{\min}(Q)\right)^2 \mathbf{E}[\|\xi_i\|_2^2] + \beta_i^2 (G^2 + cN / m^2),
\end{aligned}
\end{equation}
where $\lambda_{\min}(Q) \geq \frac{8}{9}\lambda_{\min}({A}^\intercal{C}^{-1}{A}) > 0$.
\end{lemma}


The detailed proof of Lemma \ref{lem:iteana} is in the appendix. Note that, in the stochastic programming literature, similar results rely on the assumption of a strongly convex (concave) objective function \citep{nesterov2013introductory}. However, our results show that we do not need both the primal variable, $\theta$, and dual variable, $w$, to be strongly convex (concave). This is because of the special form our objective function, i.e., our objective function is a quadratic optimization problem \citep{bertsekas1999nonlinear}.



Next, we provide the utility analysis in terms of different step size approaches. We first provide the utility bound when using a constant step size.

\begin{theorem}
\label{thm:ut1}
Let $\xi_{N + 1}$ be generated from Algorithm \ref{alg:DPEPDG}. If step size $\{\beta_i\}_{i = 1}^{\infty}$ is constant, i.e., $\beta_i = \eta/N^k < 1 / \lambda_{\min}(Q)$, where $k \in (0,1)$ and $\eta$ is any positive real number, then
\begin{align}
& \mathbf{E}[\|\xi_{N + 1}\|_2^2] \leq \\
& \underbrace{\left(1 - \frac{\eta}{N^{k}}\lambda_{\min}(Q)\right)^{2N} \left(\mathbf{E}[\|\xi_1\|_2^2] - C_0 \right)}_{\text{term \RomanNum{1}}} + \underbrace{C_0}_{\text{term \RomanNum{2}}},
\label{eq:laskjdf}
\end{align}
where $C_0 = \dfrac{\eta (G^2 + cN / m^2)}{2N^{k}\lambda_{\min}(Q) - \eta\lambda^2_{\min}(Q)}$ and $\lambda_{\min}(Q) \geq \frac{8}{9}\lambda_{\min}({A}^\intercal{C}^{-1}{A}) > 0$.
\end{theorem}

The detailed proof of Theorem \ref{thm:ut1} is in the appendix. Theorem \ref{thm:ut1} shows that there is a strong trade-off between accuracy and privacy when using a constant step size. Increasing privacy requires $c$ to become larger, which increases the right side of \eqref{eq:laskjdf}. 
Furthermore, notice that term \textit{\RomanNum{1}} has a linear rate of convergence, since we have $k \in (0,1)$, so that 
\begin{align}
& \left(1 - \frac{\eta}{N^{k}}\lambda_{\min}(Q)\right)^{2N} \\
= & \left(1 - \frac{\eta\lambda_{\min}(Q)}{N^{k}}\right)^{(N^k / (\eta\lambda_{\min}(Q))) \cdot (2\eta\lambda_{\min}(Q)N^{1 - k})} \\
\to & e^{(-2\eta\lambda_{\min}(Q)N^{1 - k})},
\end{align}
as $N$ goes to infinity, while term \textit{\RomanNum{2}} \emph{diverges} since it has $N$ in the numerator. 
Thus, initially term \textit{\RomanNum{1}} dominates and we would expect rapid convergence. However, for large $N$, term \textit{\RomanNum{2}} will eventually dominate, and the algorithm will diverge.


Next, we consider the convergence rate (utility analysis) when using a diminishing step size sequence. Theorem \ref{thm:ut2} shows that in this setting the divergent term is not present. 
\begin{theorem}
\label{thm:ut2}
Let $\xi_{N + 1}$ be generated by Algorithm \ref{alg:DPEPDG}. If $\{\beta_i\}_{i = 1}^{\infty}$ is a sequence of diminishing step sizes defined as $\beta_i = \frac{\eta}{\lambda_{\min}(Q)i}$, where $\eta > 1$, then:
\begin{align}
\mathbf{E}[\|\xi_{N + 1}\|_2^2] \leq & \frac{\max\left\{\|\xi_{1}\|_2^2, \frac{\eta^2 (G^2 + cN / m^2)}{(\eta - 1)\lambda^2_{\min}(Q)}\right\}}{N} \\
\leq &\underbrace{\frac{1}{N}\max\left\{\|\xi_{1}\|_2^2, \frac{\eta^2 G^2}{(\eta - 1)\lambda^2_{\min}(Q)}\right\}}_{\text{term \RomanNum{3}}} \\
& + \underbrace{\frac{\eta^2 c}{m^2(\eta - 1)\lambda^2_{\min}(Q)}}_{\text{term \RomanNum{4}}},
\end{align}
where $\lambda_{\min}(Q) \geq \frac{8}{9}\lambda_{\min}({A}^\intercal{C}^{-1}{A}) > 0$.
\end{theorem}





The detailed proof of Theorem \ref{thm:ut2} is in the appendix.
First, notice that Theorem \ref{thm:ut2} has the same accuracy-privacy trade-off as Theorem \ref{thm:ut1} due to its dependence on $c$. However, Theorem \ref{thm:ut2} shows that using diminishing step sizes results in a sublinear (i.e., worse than term \textit{\RomanNum{1}}) convergence rate (term \textit{\RomanNum{3}}), up to the information-theoretic limit (term \textit{\RomanNum{4}}). Since constant step sizes provide a better initial convergence rate (before term \textit{\RomanNum{2}} dominates) than diminishing step sizes, initially using a constant step size would be preferable. However, after enough iterations, the noise in the gradient prevents the constant step size algorithm from converging to an optimal solution due to term \textit{\RomanNum{2}}. Thus, in the long-term (when running many iterations), using a diminishing step size will produce a better solution. 
It should be also noted that \citet{bassily2014differentially} gave the optimal lower bound of utility for the problem of Empirical Risk Minimization (ERM) for both general convex and strongly convex loss function. Theorem 4 shows that for large $N$ (i.e., $N = O(m^2 \varepsilon^2 / d)$), our method can attain that optimal lower bound, i.e., $\Omega(d / m^2 \varepsilon^2)$.


We now consider the influence of the mini-batch size (i.e., the number of transitions used in the sampled trajectory). Let $\tau_i$ be the number of transition samples which are used in iteration $i$. The approximation error according to the definition of primal-dual gradient in \eqref{eq:pdgrad}, can be written as
$
\Delta_i = \tau_i^{-1}\sum_{t = 1}^{\tau_i}\Delta_t
$,
where $\Delta_i$ is the approximation error for only using one transition. Thus, if we replace Assumption \ref{asm3} with the assumption $\mathbf{E}[\|\Delta_i\|_2^2] \leq G^2$, then we have that:
\begin{align}
\mathbf{E}[\|\Delta_i\|_2^2] = \mathbf{E}\left[\left\|\frac{1}{\tau_i}\sum_{t = 1}^{\tau_i}\Delta_t\right\|_2^2\right] \leq \frac{G^2}{\tau_i}.
\end{align}
Thus, the variance bound is inversely proportion to the number of transitions used, and tighter variance bounds provide faster rates of convergence (as shown in Theorems \ref{thm:ut1} and \ref{thm:ut2}). Therefore, the best choice is to use all of the transitions in the $i$th trajectory when computing $B_i(\theta_i,w_i)$.



\section{Experimental Results}
\label{sec:exp}

\begin{figure*}[th]
\begin{minipage}{0.626\linewidth}
\fbox{
\begin{minipage}{\linewidth}
\begin{subfigure}{0.496\linewidth}
\includegraphics[width=\linewidth]{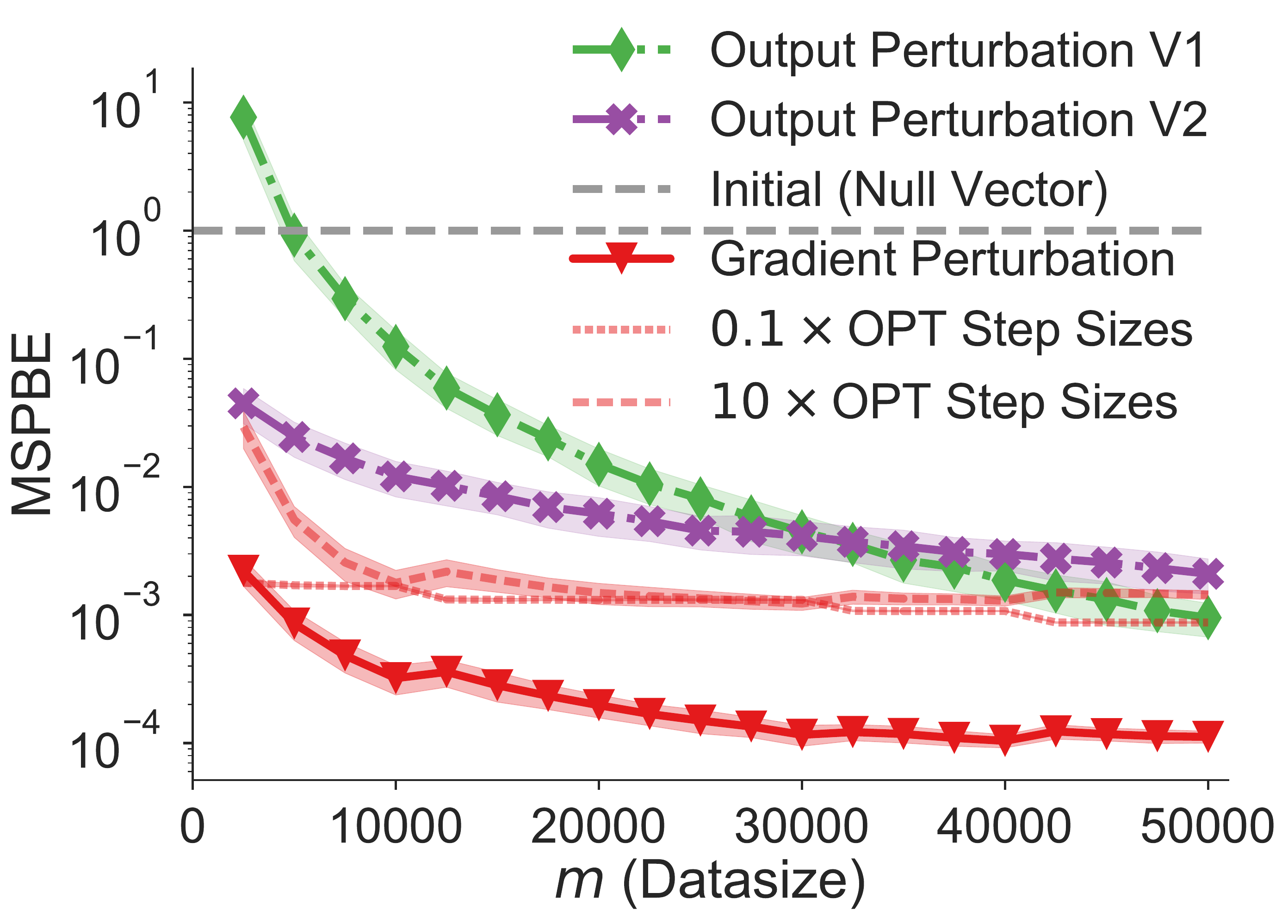}
\caption{Empirical comparison}
\label{fig:chain}
\end{subfigure}
\begin{subfigure}{0.496\linewidth}
\includegraphics[width=\linewidth]{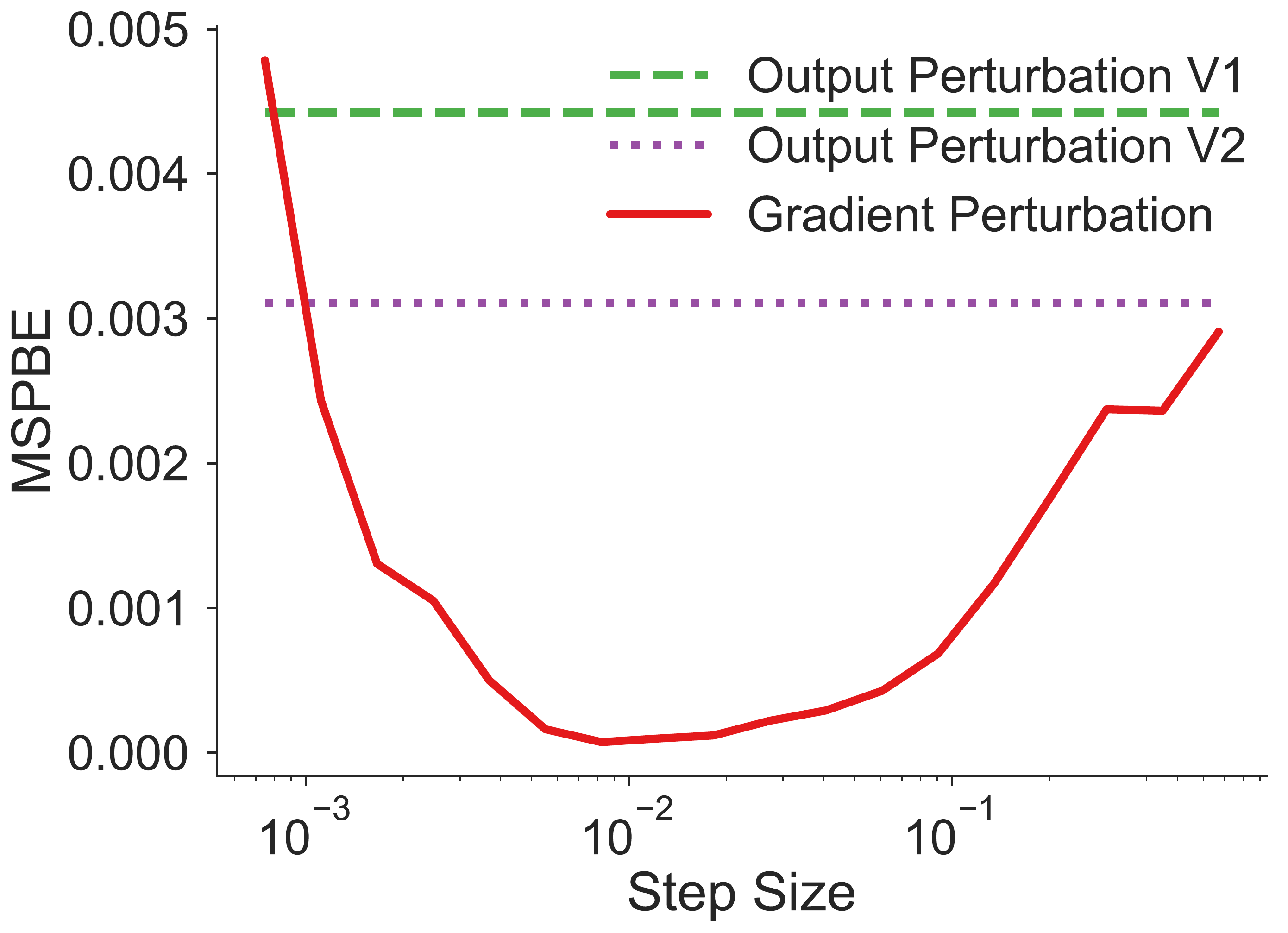}
\caption{Sensitivity of step sizes ($m = 3 \times 10^5$)}
\label{fig:chain_varistep}
\end{subfigure}
\caption{Experiments in on-policy chain}
\end{minipage}
}
\fbox{
\begin{minipage}{\linewidth}
\begin{subfigure}{0.496\linewidth}
\includegraphics[width=\linewidth]{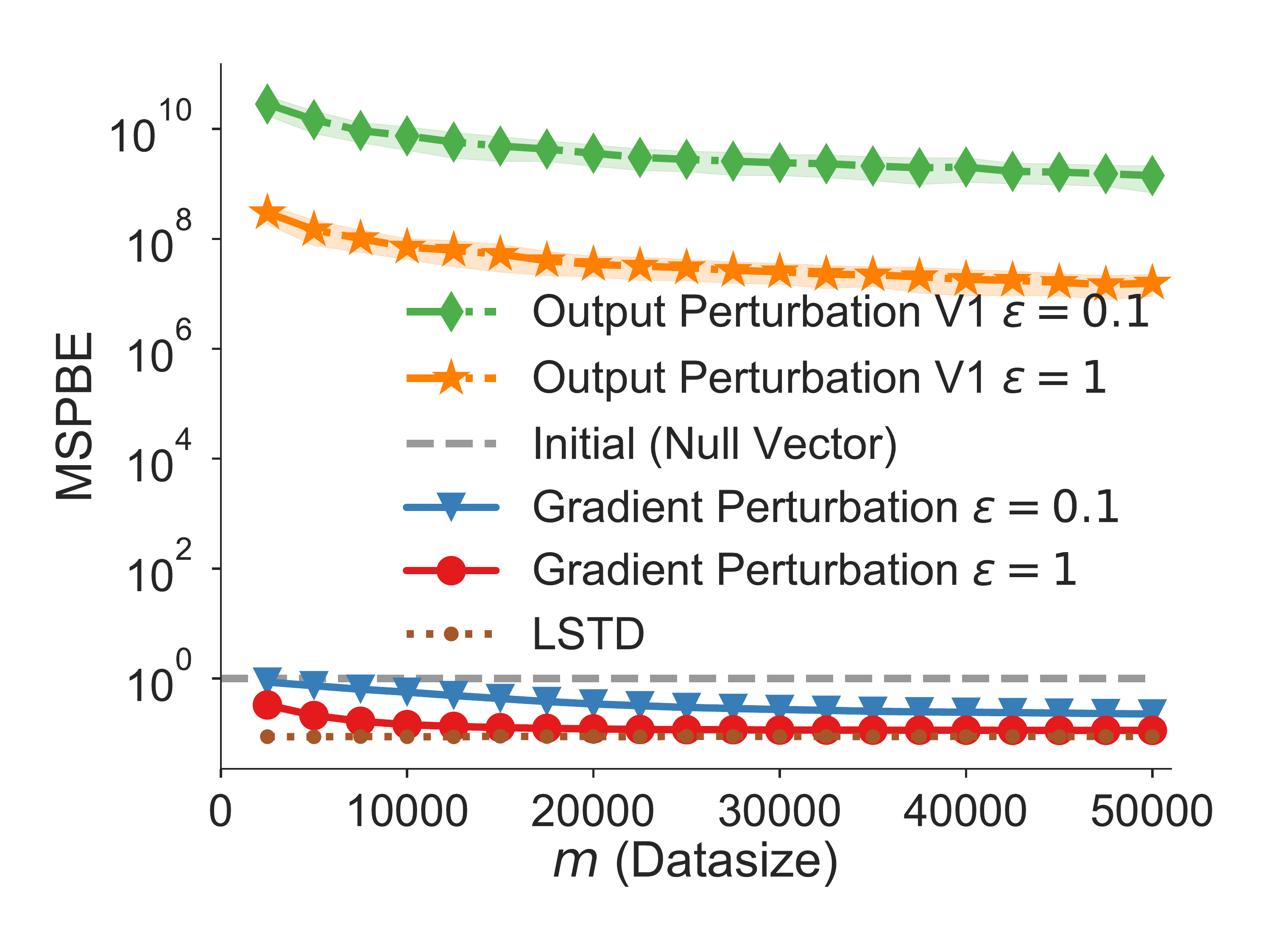}
\caption{Empirical comparison}
\label{fig:onpolicy_errbar}
\end{subfigure}
\begin{subfigure}{0.496\linewidth}
\includegraphics[width=\linewidth]{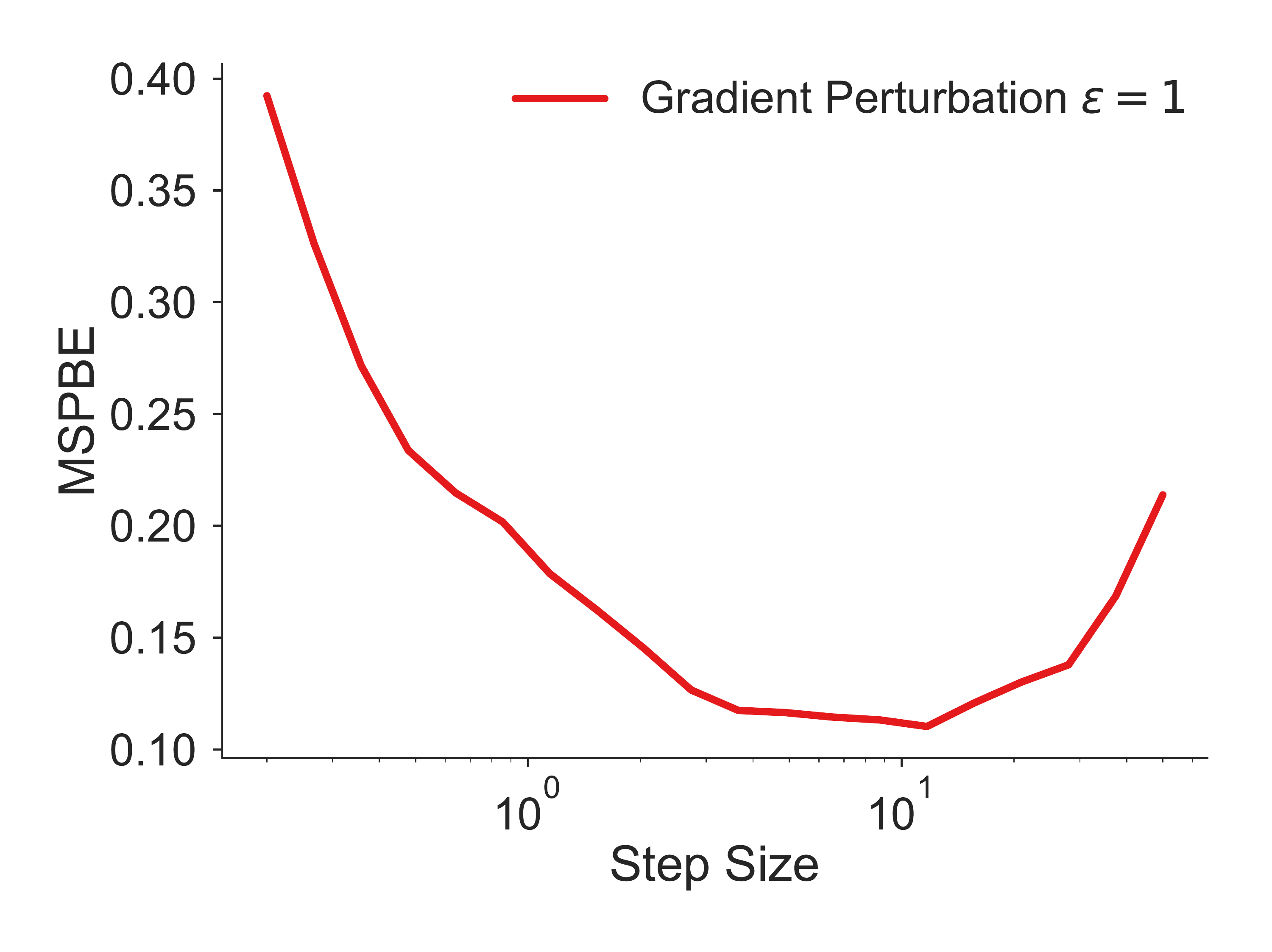}
\caption{Sensitivity of step sizes ($m = 3 \times 10^5$)}
\label{fig:on_varistep_30k}
\end{subfigure}
\caption{Experiments in on-policy mountain car}
\end{minipage}
}
\end{minipage}
\hfill
\fbox{
\begin{minipage}{0.313\linewidth}
\begin{subfigure}{\linewidth}
\includegraphics[width=\linewidth]{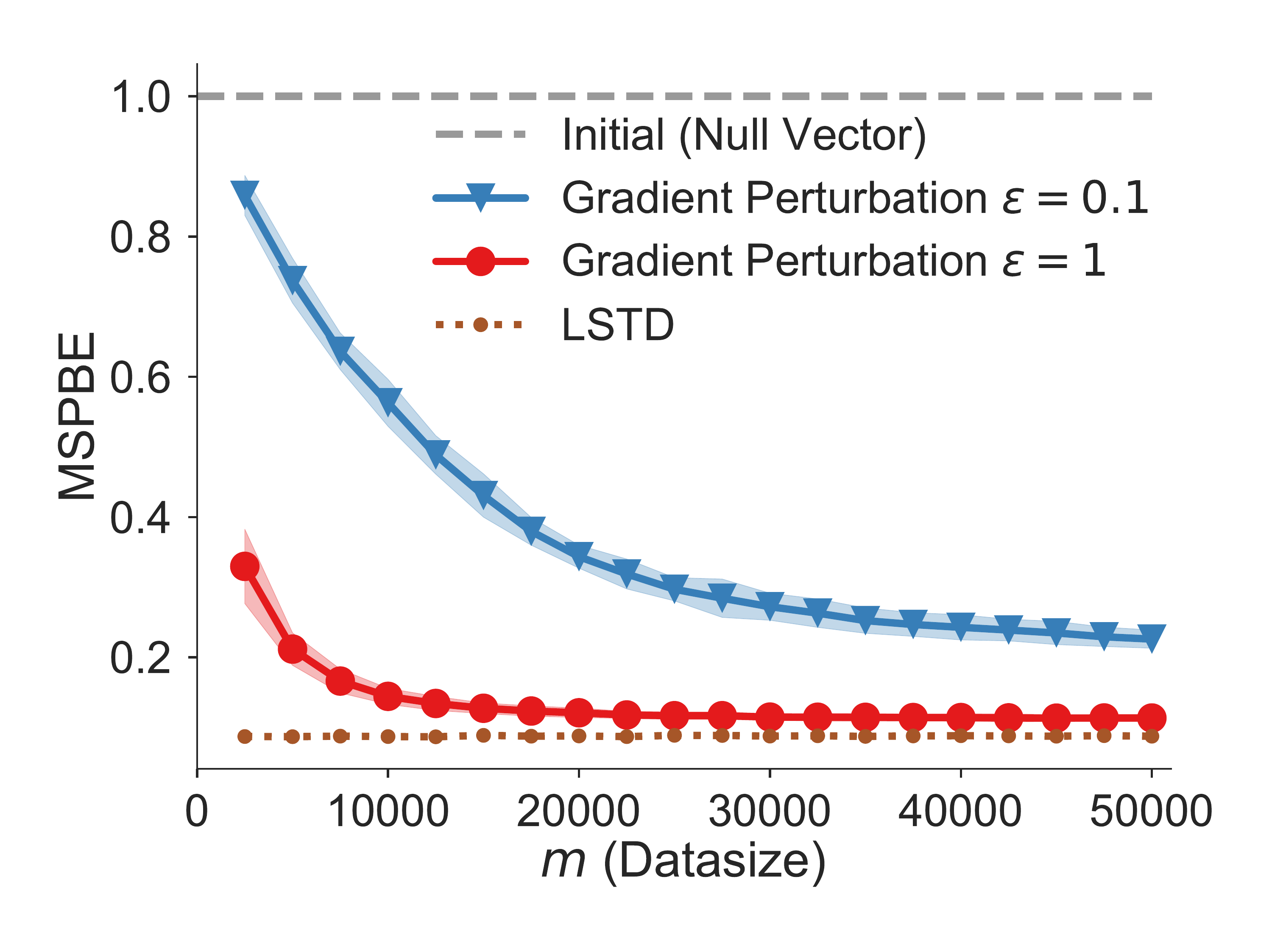}
\caption{On-policy mountain car}
\label{fig:onpolicy_only}
\end{subfigure}
\vspace{6mm}
\\
\vspace{6mm}
\begin{subfigure}{\linewidth}
\includegraphics[width=\linewidth]{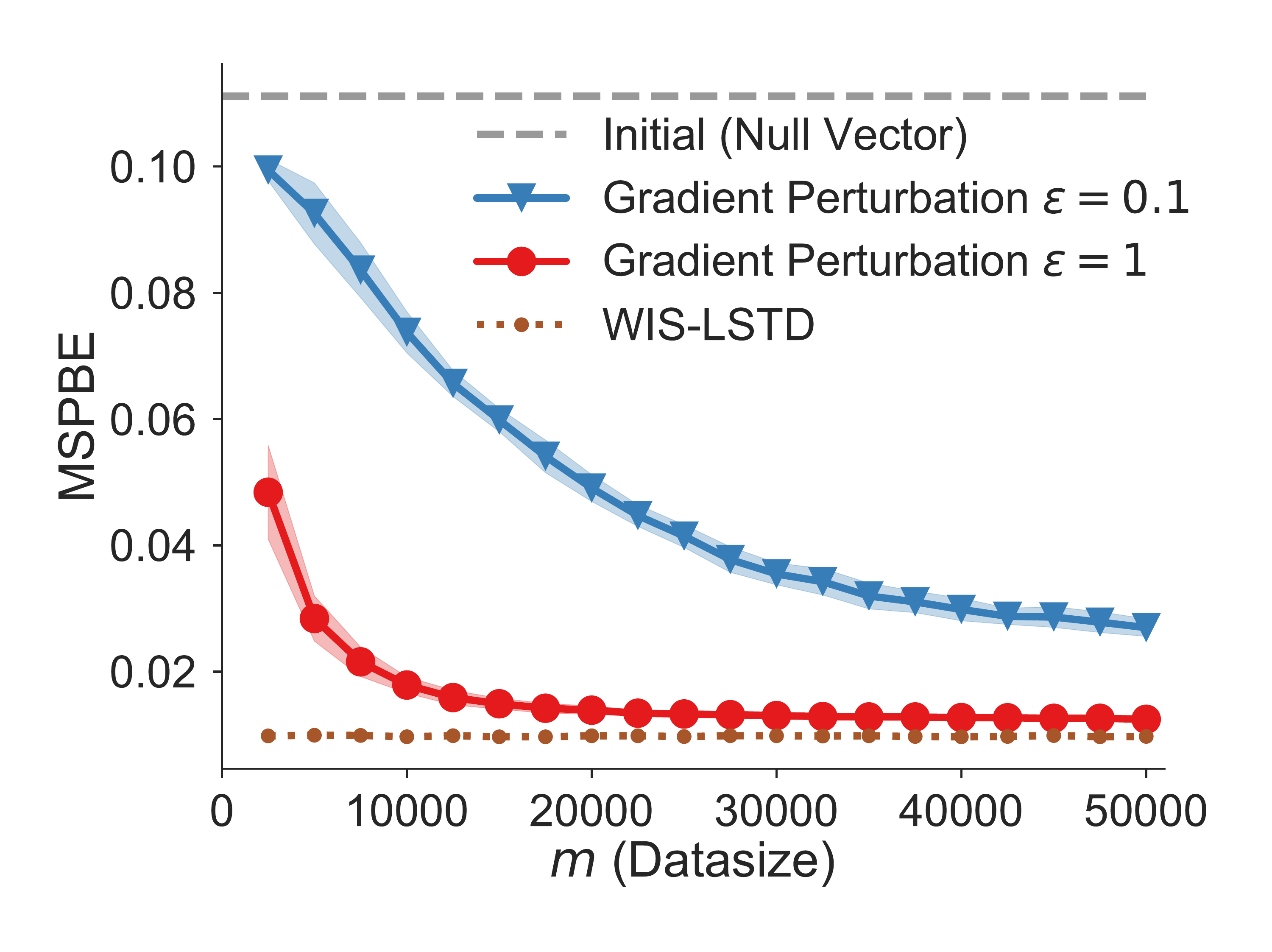}
\caption{Off-policy mountain car}
\label{fig:offpolicy}
\end{subfigure}
\caption{Detailed empirical results in mountain car}
\end{minipage}
}
\end{figure*}

In this section we compare the performance of our proposed algorithm, \emph{gradient perturbed off-policy evaluation} (GPOPE) (called gradient perturbation in this section to emphasize its difference from prior methods), with two prior methods, DP-LSW and DP-LSL \citep{balle2016differentially} on two on-policy evaluation tasks. For clarity, we use output perturbation V1 to denote DP-LSW, and output perturbation V2 to denote DP-LSL. We then illustrate the behaviour of gradient perturbation on off-policy task, a common benchmark control tasks and on a more challenging HIV simulator. 
The results we show in the following figures are all averaged over 100 trials and include standard deviation error bars, and we fix $\delta = 10^{-5}$ in all our experiments. 

\paragraph{Synthetic chain domain}

In the first on-policy task, we consider a chain domain that consists of $40$ states. The agent begins at a uniformly random state on the chain. In each state the agent has probability $p=0.5$ to stay and probability $1- p$ of advancing to the right. The agent receives a reward of $1$ when reaching the final absorbing state, and $0$ for all other states. We use  $\gamma = 0.99$, $\varepsilon = 0.1$. We compared our algorithm, gradient perturbation, with output perturbation V1 and output perturbation V2 for on-policy evaluation in the tabular setting. This toy example illustrates one typical case in medical applications \citep{balle2016differentially}, where patients tend to progress through stages of recovery at different speeds, and past states are not typically revisited (partly because in the medical domain, states contain historical information about past treatments). The main result is shown in Figure \ref{fig:chain}, where MSPBE denotes \emph{mean squared projected Bellman error} (a common measure of inaccuracy for policy evaluation in reinforcement learning \citep{sutton2009fast}), and where the datasize, $m$, is the number of trajectories used.

We use different step sizes (a hyper-parameter) for different $m$ (amounts of data). Since the choice of step size cannot depend on the private data (this choice could leak information not captured by our analysis), we assume that the step size was tuned using similar public data---a common approach in differential privacy \citep{papernot2016semi}. 
For Figure \ref{fig:chain}, we assume that this method was used to obtain optimal step sizes. Our proposed method outperforms output perturbation V1 and output perturbation V2 in terms of accuracy by an order of magnitude.


In practice there may not always be public data similar to the private data, or the public data may differ slightly from the private data. This means that the optimal step size for the public data may not be the optimal step size for the private data. Therefore, it is necessary to test the robustness of our algorithm to changing hyper-parameters. Since step size is the only variable hyper-parameter for different $m$, Figure \ref{fig:chain} also shows the results of using $0.1 \times$ and $10 \times$ the optimal step sizes. Even using these imprecise optimal step sizes, our proposed approach usually achieves better accuracy than prior methods. Figure \ref{fig:chain_varistep} shows the accuracy when the step size varies, but the amount of data, $m$, is fixed. This shows that accuracy is stable for a very wide range of step sizes. We provide additional experiments in the appendix to further show the robustness of our algorithm to the step size parameter.


\paragraph{Mountain Car}

Next we performed these same experiments using the mountain car domain \citep{sutton1998reinforcement} for on-policy policy evaluation. Mountain car is a popular RL benchmark problem with a two dimensional continuous state space, three discrete actions, and deterministic dynamics. We first used Q-learning with the fifth order Fourier basis \citep{konidaris2011value} to obtain a decent policy to evaluate. We ran this policy to collect the trajectories that comprise the data set, and used our gradient perturbation algorithm and the output perturbation algorithms to estimate the value function for the learned policy. 


Figure \ref{fig:onpolicy_errbar} shows the accuracy of our algorithm and compares with output perturbation V1 and \textit{least squares temporal difference} \citep[LSTD]{LSTD}. LSTD does not provide any privacy guarantees, and is presented here to show how close our algorithm is in accuracy to non-private methods. Note that output perturbation V2 fails to guarantee differential privacy for MDPs with continuous states or actions. 
While Figure \ref{fig:onpolicy_errbar} shows that our proposed gradient perturbation algorithm improves upon existing methods by orders of magnitude, Figure \ref{fig:onpolicy_only} provides a zoomed in view of the same plot to show the speed with which our algorithm converges when using different privacy settings. 
%
Similar to the chain domain, we show the robustness of our algorithm to step sizes in Figure \ref{fig:on_varistep_30k}, and present additional experiments in the appendix.


We also tested our algorithm on the mountain car domain for \emph{off-policy} evaluation. Since LSTD is an on-policy algorithm, here we compare to a non-private off-policy variant of LSTD, called WIS-LSTD \citep{Mahmood2014}. Note that output perturbation methods fail to guarantee differential privacy for off-policy evaluation, and so we only evaluate our algorithm in this part. Figure \ref{fig:offpolicy} shows the result of gradient perturbation for off-policy evaluation for the mountain car domain with different privacy settings. The behavior policy, $\pi_b$, is the policy learned by Q-learning, and the evaluated policy is the uniform policy. Despite being off-policy (which usually increases data requirements relative to on-policy problems), our algorithm's performances in Figures \ref{fig:onpolicy_only} and \ref{fig:offpolicy} are  remarkably similar. 

\paragraph{HIV simulator} We also evaluate our approach on an HIV treatment simulation domain. This simulator was first introduced by \citet{ernst2006clinical}, and consists of six features describing the state of the patient and four possible actions. Compared with the two domains above, this simulator is much closer to the practical medical treatment design, and its dynamics are more complex. 

\begin{figure}[htb]
\centering
\includegraphics[width=0.35\linewidth]{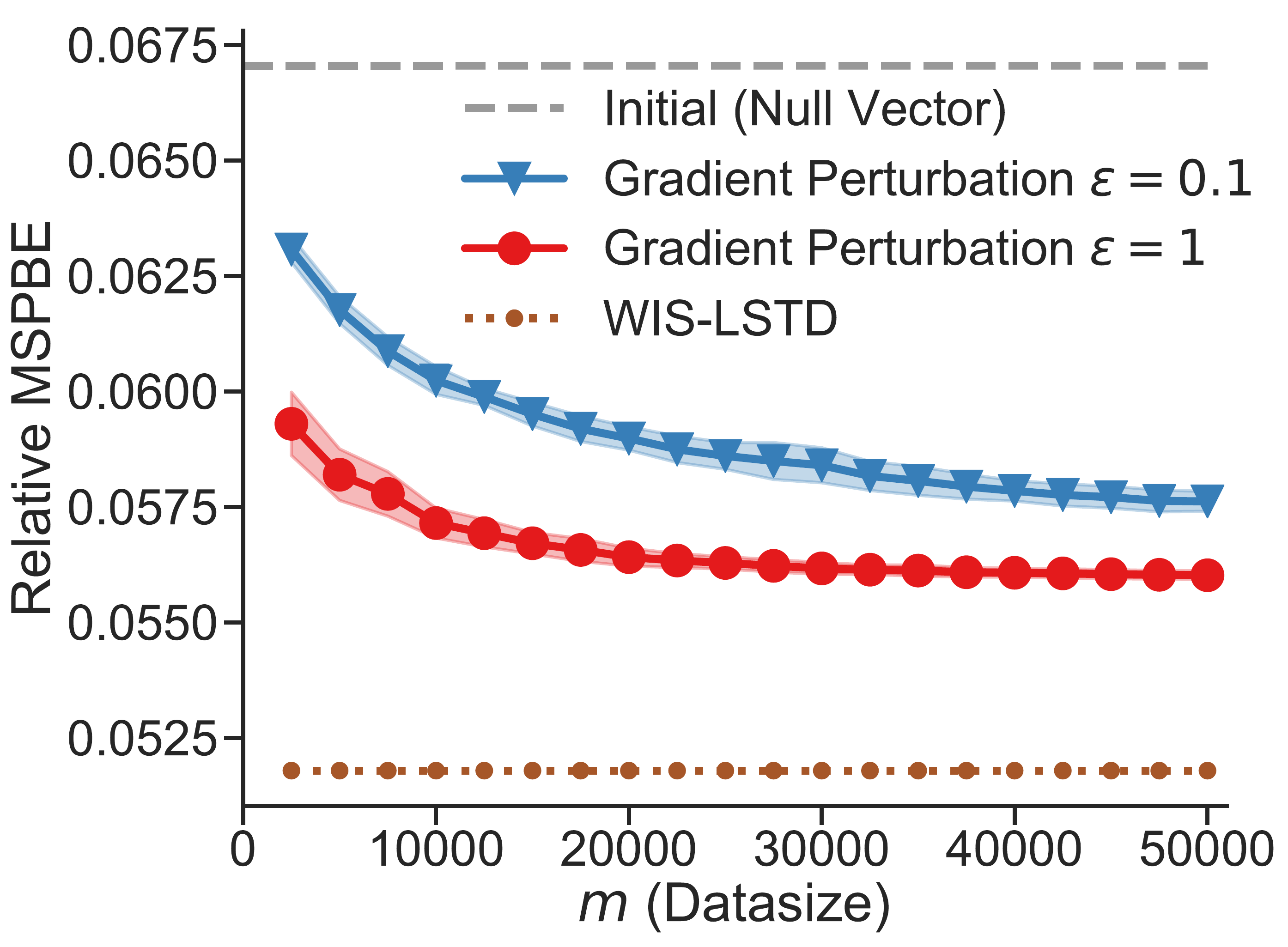}
\caption{Off-policy HIV domain}
\label{fig:offpolicy_HIV}
\end{figure}

Figure \ref{fig:offpolicy_HIV} shows the results on the HIV simulator. We obtain the policy that is evaluated using Q-learning, and use a policy that is softmax w.r.t.~the optimal Q function as the behavior policy. We use relative MSPBE in Figure \ref{fig:offpolicy_HIV}, which normalizes MSPBE using the average reward ($\sim 10^5$) of the evaluated policy.


\section{Discussion and Conclusion}
\label{sec:discussion}

To protect individual privacy when applying reinforcement learning algorithms to sensitive training data, we present the first differentially private algorithm for off-policy evaluation. Our approach extends on the TD methods and comes with a privacy analysis and a utility (convergence rate) analysis. The utility guarantee shows that the privacy cost can be diminished by increasing the size of training batches, and the privacy/utility trade-off can be optimized by using a decaying step size sequence. In our experiments, our algorithm, gradient Perturbed off-policy evaluation (GPOPE), outperforms the previous methods in the restricted on-policy setting that prior work considers, can work well for both discrete and continuous domains, and guarantees differential privacy for both on-policy and off-policy evaluation problems. 
We also demonstrate the effectiveness of our approach in both common benchmark tasks and on a more challenging HIV simulator.
Since our approach is based on gradient computations, it can be extended easily to more advanced first-order optimization methods, such as stochastic variance reduction methods \citep{du2017stochastic,palaniappan2016stochastic}, and momentum methods \citep{nesterov2013introductory}.




\bibliography{ref}
\bibliographystyle{apalike}

\clearpage

\appendix
\onecolumn


\begin{center}
{\LARGE \textbf{Appendix}}
\end{center}

\section{Relationships Among Parameters}

Table \ref{tb:relationship} shows the relationships among privacy parameters $\varepsilon,\delta,\sigma$, the total number of iterations $N$, and the size of dataset $m$. We use the color of red to denote negatively related, green to denote positively related. For example, if $\varepsilon$ is decreased, and only $\delta$ is changed, then $\delta$ must be increased. Similarly, if size of dataset $m$ is increased, and only $\varepsilon$ changed, then $\varepsilon$ must be decreased.

\begin{table}[htbp]
\centering
\begin{tabular}{|c|c|c|c|c|c|}
\hline
              & $\varepsilon$ & $\delta$ & $N$ & $\sigma$ & $m$ \\ \hline
$\varepsilon$ & \cellcolor{gray!50}  & \cellcolor{red!50} & \cellcolor{green!50} & \cellcolor{red!50} & \cellcolor{red!50}        \\ \hline
$\delta$      & \cellcolor{red!50}  & \cellcolor{gray!50} & \cellcolor{green!50} & \cellcolor{red!50} & \cellcolor{red!50}       \\ \hline
$N$           & \cellcolor{green!50}  & \cellcolor{green!50} & \cellcolor{gray!50} & \cellcolor{green!50} & \cellcolor{green!50}       \\ \hline
$\sigma$      & \cellcolor{red!50}  & \cellcolor{red!50} & \cellcolor{green!50} & \cellcolor{gray!50}  & \cellcolor{red!50}      \\ \hline
$m$      & \cellcolor{red!50}  & \cellcolor{red!50} & \cellcolor{green!50} & \cellcolor{red!50}  & \cellcolor{gray!50}      \\ \hline
\end{tabular}
\caption{Relationship Matrix among $\varepsilon$, $\delta$, $N$, $\sigma$, $m$ (red denotes negatively related, green denotes positively related)}
\label{tb:relationship}
\end{table}

\section{Proofs in Privacy Analysis}
\label{sec:privana_app}

In this section we provide a detailed analysis of the privacy guarantee of our algorithm. We first introduce the following key definitions and properties we will use.

\begin{definition}[R\'enyi Divergence]
Let $P$ and $Q$ be probability distributions on $\omega$. For $\alpha \in (1,\infty)$, we define the R\'enyi divergence of order $\alpha$ between $P$ and $Q$ as
\begin{align}
D_{\alpha}(P \| Q) = & \frac{1}{\alpha - 1}\log\left( \int_{\Omega} P(x)^\alpha Q(x)^{1 - \alpha} \mathrm d x \right) \\
= & \frac{1}{\alpha - 1} \log\left( \mathbf E_{x \sim Q} \left[ \left( \frac{P(x)}{Q(x)} \right)^\alpha \right] \right) \\
= & \frac{1}{\alpha - 1} \log\left( \mathbf E_{x \sim P} \left[ \left( \frac{P(x)}{Q(x)} \right)^{\alpha - 1} \right] \right),
\end{align}
where $P(\cdot)$ and $Q(\cdot)$ are the probability density functions of $P$ and $Q$ respectively.
\end{definition}

\begin{definition}[R\'enyi Differential Privacy]
We say that a mechanism $\mathcal M$ is $(\alpha,\varepsilon)$-R\'enyi Differential Privacy (RDP) with order $\alpha \in (1,\infty)$ if for all neighboring dataset $d,d'$
\begin{align}
D_{\alpha}(\mathcal M(d) \| \mathcal M(d')) \coloneqq \frac{1}{\alpha - 1} \log\left( \mathbf E_{x \sim \mathcal M(d')(x)} \left[ \left( \frac{\mathcal M(d)(x)}{\mathcal M(d')(x)} \right)^\alpha \right] \right) \leq \varepsilon.
\end{align}

\end{definition}

\begin{lemma}[Lemma 2.5 in \citep{bun2016concentrated}]
\label{lem:renyidiv}
Let $\nu, \mu \in \mathbb R^d$, $\sigma \in \mathbb R$, and $\alpha \in [1, \infty)$. Then, 
\begin{align}
D_{\alpha} (\mathcal N(\mu, \sigma I_d) \| \mathcal N(\nu, \sigma I_d)) = \frac{\alpha \|\mu - \nu\|_2^2}{2 \sigma^2}.
\end{align}
\end{lemma}



\subsection{Proof of Lemma \ref{lem:privlem}}
\label{subsec:proprivlem}
\begin{proof}

Let fixed $d'$ and let $d = d' \cup x_m$, where $x_m$ denotes trajectory $m$ with length ${\tau_m}$. Without loss of generality, let $\bar{B}_t(x_m) =  \mathbf{e_1}$ and $\sum_{x_i \in d'} \bar{B}_t(x_m) = \mathbf{0}$. Thus $\mathcal{M}(d)$ and $\mathcal{M}(d')$ are distributed identically except for the first coordinate. Hence we transfer it to a one-dimension problem. Let $\mu_0$ denote the probability density function of $\mathcal{N}(0,\sigma^2)$ and let $\mu_1$ denote probability density function of $\mathcal{N}(1,\sigma^2)$. Thus,
\begin{align}
\mathcal{M}(d) &\sim \mu_0, \\
\mathcal{M}(d') &\sim \mu \coloneqq \left\{\begin{array}{cl}
    \mu_0, & \text{w.p.}~~ 1 - q  \\
    \mu_1, & \text{w.p.}~~ q
\end{array}\right.,
\end{align}
where $q = 1 / m$. To avoid the difficulty of analysis this complex mixture distribution, we decompose $\mathcal M$ as a composition of two algorithm $\mathcal M_0 \circ \texttt{subsample}$ which is defined as: (1) \texttt{subsample}: subsample without replacement 1 datapoint of the dataset, and (2) a randomized algorithm taking the subsampled dataset as the input. Next, we use the amplification properties for RDP via subsampling to obtain $\alpha_{\mathcal M}(\lambda)$.

First, we analysis the RDP for $\mathcal M_0$ as: 
\begin{align}
\varepsilon_0(\alpha) \coloneqq & D_{\alpha}(\mathcal M_0 (d) \| \mathcal M_0(d')) \\
= & D_{\alpha}(\mu_0 \| \mathcal \mu_1) \\
= & \frac{\alpha}{2 \sigma^2},
\label{eq:rdpbd0}
\end{align}
where the last equation follows from Lemma \ref{lem:renyidiv}.

By the amplification properties for RDP via subsampling (Theorem 9 in \citep{wang2018subsampled}), we can obtain $\mathcal M = \mathcal M_0 \circ \texttt{subsample}$ is ($\alpha, \varepsilon(\alpha)$)-RDP,
\begin{align}
\label{eq:rdpbd}
\varepsilon(\alpha) \leq \frac{1}{\alpha - 1} \log\left(1 + \frac{\alpha(\alpha - 1)}{2 m^2} \min \left\{ 4\left(e^{\varepsilon_0(2)} - 1\right), e^{\varepsilon_0(2)} \min\left\{2,\left(e^{\varepsilon_0(\infty)} - 1\right)^2\right\}  \right\} \right),
\end{align}
where $\varepsilon_0(\alpha)$ is defined in \eqref{eq:rdpbd0}, and we ignored the higher-order terms since $m \gg 1$. According to the definition of RDP, we have 
\begin{align}
D_{\alpha}(\mathcal M (d) \| \mathcal M(d')) \leq \varepsilon(\alpha).
\label{eq:div_edp}
\end{align}

Since the Gaussian mechanism does not have a bound $\varepsilon_0(\infty)$, term $\min \left\{ 4\left(e^{\varepsilon_0(2)} - 1\right), e^{\varepsilon_0(2)} \min\left\{2,\left(e^{\varepsilon_0(\infty)} - 1\right)^2\right\}  \right\}$ in the bound \eqref{eq:rdpbd} can be simplified as $\min \left\{ 4\left(e^{\varepsilon_0(2)} - 1\right), 2 e^{\varepsilon_0(2)}  \right\}$, where $\varepsilon_0(2) = 1 / \sigma^2$ according to \eqref{eq:rdpbd0}.


By properties of R\'enyi divergence, we have
\begin{align}
\alpha_{\mathcal M}(\lambda) = & \lambda D_{\lambda + 1}(\mathcal{M}(d) \| \mathcal{M}(d')) \\
\leq & \lambda \varepsilon(\lambda + 1) \\
\leq & \lambda \frac{1}{\lambda} \log\left(1 + \frac{\alpha(\alpha - 1)}{2 m^2} \min \left\{ 4\left(e^{\varepsilon_0(2)} - 1\right), 2 e^{\varepsilon_0(2)}  \right\} \right) \\
= & \log\left(1 + \frac{\lambda(\lambda + 1)}{2 m^2} \min \left\{ 4\left(e^{1 / \sigma^2} - 1\right), 2 e^{1 / \sigma^2}  \right\} \right) \\
\leq & \frac{\lambda(\lambda + 1)}{2 m^2} \min \left\{ 4\left(e^{1 / \sigma^2} - 1\right), 2 e^{1 / \sigma^2}  \right\},
\end{align}
where the second inequality follows from \eqref{eq:div_edp}, the third inequality follow from \eqref{eq:rdpbd}. This completes the proof.

\end{proof}

\subsection{Proof of Theorem \ref{thmdp}}
\label{subsec:provprivthm}

We first introduce a useful theorem for the calculation of 

\begin{theorem} [Theorem 2 in \citep{abadi2016deep}]
\label{thm1}
Let $\alpha_{\mathcal{M}}(\lambda)$ be the moments accountant of a randomized mechanism $\mathcal{M}$. 
\begin{enumerate}
    \item \textbf{[Composability]} Suppose that a mechanism $\mathcal{M}$ consists of a sequence of adaptive mechanisms $\mathcal{M}_{1}$, $\ldots$, $\mathcal{M}_{k}$ where
    %
    %
    $\mathcal{M}_{i}$:  $\prod_{j = 1}^{i - 1}\mathcal{Y}_{j}\times\mathcal{D}\rightarrow \mathcal{Y}_{i}$. Then, for any $\lambda$
    \begin{equation}
        \alpha_{\mathcal{M}}(\lambda) \leq \sum_{i = 1}^{k}\alpha_{\mathcal{M}_{i}}(\lambda).
    \end{equation}
    \item \textbf{[Tail bound]} For any $\varepsilon > 0$, the mechanism $\mathcal{M}$ is $(\varepsilon, \delta)$-differentially private for
    \begin{equation}
        \delta = \min_{\lambda}\exp(\alpha_{\mathcal{M}}(\lambda) - \lambda\varepsilon).
    \end{equation}
\end{enumerate}
\end{theorem}

Theorem \ref{thm1} enables us to compute and bound the moments accountant, $\alpha_{\mathcal{M}}(\lambda)$, at each iteration and sum them to bound the moments of the whole algorithm. This allows us to convert the moments bound to the $(\epsilon,\delta)$-differential privacy guarantee.

Given Lemma \ref{lem:privlem} and Theorem \ref{thm1}, the proof that Algorithm \ref{alg:DPEPDG} is $(\varepsilon,\delta)$-differential private can be obtained directly, because Lemma \ref{lem:privlem} bounds the moments of each iteration, and we can calculate the moments accountant of our whole algorithm by applying Theorem \ref{thm1}. The proof of Theorem \ref{thmdp} is as follows.

\begin{proof}

We first analysis the term $\left\{ 4\left(e^{1 / \sigma^2} - 1\right), 2 e^{1 / \sigma^2}  \right\}$ in Lemma \ref{lem:privlem}. If $\sigma^2 \geq 1 / \ln 2$, we have
\begin{align}
& \min \left\{ 4\left(e^{\varepsilon_0(2)} - 1\right), 2 e^{\varepsilon_0(2)}  \right\} \\
= & \min \left\{ 4\left(e^{1 / \sigma^2} - 1\right), 2 e^{1 / \sigma^2}  \right\} \\
= & 4\left(e^{1 / \sigma^2} - 1\right) \\
\leq & \frac{8}{\sigma^2},
\end{align}
and
\begin{align}
\alpha_{\mathcal M}(\lambda) \leq \frac{4 \lambda(\lambda + 1)}{m^2 \sigma^2}.
\end{align}
If $\sigma^2 < 1 / \ln 2$, we have
\begin{align}
& \min \left\{ 4\left(e^{\varepsilon_0(2)} - 1\right), 2 e^{\varepsilon_0(2)}  \right\} \\
= & \min \left\{ 4\left(e^{1 / \sigma^2} - 1\right), 2 e^{1 / \sigma^2}  \right\} \\
= & 2 e^{1 / \sigma^2},
\end{align}
and
\begin{align}
\alpha_{\mathcal M}(\lambda) \leq \frac{\lambda(\lambda + 1) e^{1 / \sigma^2}}{m^2}.
\end{align}

By Theorem \ref{thm1} and Lemma \ref{lem:privlem}, the $\log$ moment of Algorithm \ref{alg:DPEPDG} can be bounded as
$
\alpha(\lambda) \leq 4 N\lambda^2 / (m^2 \sigma^2).
$ (assuming we set $\sigma$ explicitly to satisfy $\sigma \geq \sqrt{1 / \ln 2} \approx 1.201$).
In order to use Theorem \ref{thm1} to guarantee the $(\varepsilon,\delta)$-differential privacy of Algorithm \ref{alg:DPEPDG}, we need $\lambda$ satisfy
\begin{align}
\lambda \leq \sigma^2\log\left(\frac{m}{\sigma}\right)
\end{align}
and $\sigma$ to satisfy
\begin{align}
\delta = \min_{\lambda}\exp\left(\frac{N\lambda^2}{m^2 \sigma^2} - \lambda\varepsilon\right) \leq \exp\left( - \frac{m^2 \sigma^2 \varepsilon^2}{4 N}\right).
\end{align}

Thus, when $\varepsilon = c_1 N / m^2$, all these conditions are satisfied by setting
\begin{align}
\sigma = \frac{c_2\sqrt{N \log(1/\delta)}}{m \varepsilon},
\end{align}
for some explicit constants $c_1$ and $c_2$.
\end{proof}

\section{Proofs in Utility Analysis}
\label{sec:utiana_app}

In this section we provide detailed proofs of utility analysis. First, we derive properties of each iteration of our algorithm. We assume that all transitions in the sampled trajectory are used in this subsection (as in the GPOPE algorithm).

We first provide the proof of lemma \ref{lem:iteana}.

\subsection{Proof of Lemma \ref{lem:iteana}}

\begin{proof}
The updates have the following iteration
\begin{align}
& \left[
 \begin{matrix}
   \theta_{i + 1} \\
   w_{i + 1}
  \end{matrix}
  \right] \\
  =  & \left[
 \begin{matrix}
   \theta_{i} \\
   w_{i}
  \end{matrix}
  \right]-
\beta_i B_i(\theta_i,w_i), \\
= & \left[
 \begin{matrix}
   \theta_{i} \\
   w_{i}
  \end{matrix}
  \right]-
\beta_i B(\theta_i,w_i) + \beta_i \Delta_i, \\
=  & \left[
 \begin{matrix}
   \theta_{i} \\
   w_{i}
  \end{matrix}
  \right]-
\beta_i
  \left(
   \left[
 \begin{matrix}
   0        & -{A}^\intercal\\
   {A}  & {C}
  \end{matrix}
  \right]
  \left[
 \begin{matrix}
   \theta_{i} \\
   w_{i}
  \end{matrix}
  \right]-
  \left[
 \begin{matrix}
   0 \\
   {b}
  \end{matrix}
  \right]
  \right) + \beta_i \Delta_i.
\end{align}
%
Subtracting optimal solution $(\theta^*,w^*)$ (defined in \eqref{eq:optimal}) from both sides and using the first order optimally condition, we obtain
\begin{align}
& \left[
 \begin{matrix}
   \theta_{i + 1} - \theta^*   \\
   w_{i + 1} - w^*
  \end{matrix}
  \right] \\
=  &\left[
 \begin{matrix}
   \theta_{t} - \theta^*\\
   w_{t} - w^*
  \end{matrix}
  \right]-
\beta_i
   \left[
 \begin{matrix}
   0        & -{A}^\intercal\\
   {A}  & {C}
  \end{matrix}
  \right]
  \left[
 \begin{matrix}
   \theta_{t} - \theta^* \\
   w_{t} - w^*
  \end{matrix}
  \right] + \beta_i \Delta_i.
\end{align}

The analysis of the convergence rate examines the difference between the current parameters and the optimal solution.
Note the residual vector $\xi_i$ in \eqref{eq:resvec_Q}, obeys the following iteration:
\begin{align}
\label{eq:updatelaw}
\xi_{i + 1} = (I - \beta_i Q)\xi_i + \beta_i \Delta_i,
\end{align}
where $Q$ is also defined in \eqref{eq:resvec_Q}.
Taking the Euclidean norm of each side of Eq.~\eqref{eq:updatelaw}, we obtain
\begin{align}
& \|\xi_{i + 1}\|_2^2 \\
= & \| (I - \beta_i Q) \xi_i\|_2^2 + \beta_i^2 \|\Delta_i\|_2^2 + 2\langle (I - \beta_i Q)\xi_i, \beta_i \Delta_i\rangle,
\label{eq:norm}
\end{align}
which follows from the rule that $\| a+b\|_2^2=\| a \|_2^2 + \| b \|_2^2 + 2 \langle a,b\rangle$.
Taking the expectation of both sides of Eq.~\eqref{eq:norm}, we obtain
\begin{align}
& \mathbf{E}[\|\xi_{i + 1}\|_2^2] \\
\overset{\text{(a)}}{=} &\mathbf{E}[\|(I - \beta_i Q) \xi_i\|_2^2] + \beta_i^2 \mathbf{E} [\|\Delta_i\|_2^2]\\
\overset{\text{(b)}}{\leq} & \mathbf{E}[\|I - \beta_i Q\|_{S}^2\|\xi_i\|_2^2] + \beta_i^2 \mathbf{E} [\|\Delta_i\|_2^2]\\
= & \|I - \beta_i Q\|_{S}^2 \mathbf{E}[\|\xi_i\|_2^2] + \beta_i^2 \mathbf{E} [\|\Delta_i\|_2^2]
\label{eq:expnorm}
\end{align}
where $\|I - \beta_i Q\|_{S}$ denotes the spectral norm of $(I - \beta_i Q)$, i.e., the square root of the maximum eigenvalue of $(I - \beta_i Q)$, and where {\bf (a)} holds because $\mathbf{E}[\Delta_i] = 0$, {\bf (b)} holds by a property of spectral norm \citep{meyer2000matrix}.


In order to obtain $\|I - \beta_i Q\|_{S}$, we calculate the maximum eigenvalue of $Q$ (we use $\lambda_{\max}(\cdot)$ to denote maximum eigenvalue), and the minimum eigenvalue of $Q$ (we use $\lambda_{\min}(\cdot)$ to denote minimum eigenvalue). Using the eigen-analysis of $Q$ in the previous work, in \citep{du2017stochastic}, Appendix A.3, we have
\begin{align}
\lambda_{\max}(Q) \leq & 9\kappa({C}) \lambda_{\max}({A}^\intercal{C}^{-1}{A}),\\
\lambda_{\max}(Q) \geq & \frac{8}{9}\lambda_{\min}({A}^\intercal{C}^{-1}{A}) > 0,
\end{align}
where we use $\kappa(\cdot)$ to denote $\lambda_{\max}(\cdot) / \lambda_{\min}(\cdot)$.

Choosing $\beta_i \leq 1 / \lambda_{\max}(Q)$, then we have that
\begin{align}
\label{eq:specnorm}
\|I - \beta_\theta Q\|_{S}^2 = & \left(1 - \beta_i\lambda_{\min}(Q)\right)^2.
\end{align}

Substituting \eqref{eq:specnorm} into \eqref{eq:expnorm},
\begin{align}
& \mathbf{E}[\|\xi_{i + 1}\|_2^2] \\
\leq & \left(1 - \beta_i\lambda_{\min}(Q)\right)^2 \mathbf{E}[\|\xi_i\|_2^2] + \beta_i^2 \mathbf{E} [\|\Delta_i\|_2^2] \\
\leq & \left(1 - \beta_i\lambda_{\min}(Q)\right)^2 \mathbf{E}[\|\xi_i\|_2^2] + \beta_i^2 (G^2 + cN / m^2),
\end{align}
where the second inequality follows from the assumption of variance bound in Assumption \ref{asm3}.
\end{proof}

We now prove the utility theorems using lemma \ref{lem:iteana}.

\subsection{Proof of Theorem \ref{thm:ut1}}
\label{subsec:pro_thmut1}

\begin{proof}
Let $\beta_i = \beta$. Then \eqref{eq:finalineq} leads to
\begin{align}
&\mathbf{E}[\|\xi_{i + 1}\|_2^2] - \frac{\beta (G^2 + cN / m^2)}{2\lambda_{\min}(Q) - \beta\lambda^2_{\min}(Q)}  \\
\leq & \left(1 - \beta\lambda_{\min}(Q)\right)^2 \mathbf{E}[\|\xi_i\|_2^2] + \beta^2 (G^2 + cN / m^2) - \frac{\beta (G^2 + cN / m^2)}{2\lambda_{\min}(Q) - \beta\lambda^2_{\min}(Q)}  \\
= & \left(1 - \beta\lambda_{\min}(Q)\right)^2 \left(\mathbf{E}[\|\xi_i\|_2^2] - \frac{\beta (G^2 + cN / m^2)}{2\lambda_{\min}(Q) - \beta\lambda^2_{\min}(Q)} \right).
\end{align}

Thus, recursively we have
\begin{align}
& \mathbf{E}[\|\xi_{N + 1}\|_2^2] - \frac{\beta (G^2 + cN / m^2)}{2\lambda_{\min}(Q) - \beta\lambda^2_{\min}(Q)} \\
\leq & \left(1 - \beta\lambda_{\min}(Q)\right)^{2N} \left(\mathbf{E}[\|\xi_1\|_2^2] - \frac{\beta (G^2 + cN / m^2)}{2\lambda_{\min}(Q) - \beta\lambda^2_{\min}(Q)} \right) \mathbf{E}[\|\xi_{N + 1}\|_2^2] \\
\leq & \left(1 - \beta\lambda_{\min}(Q)\right)^{2N} \cdot\left(\mathbf{E}[\|\xi_1\|_2^2] - \frac{\beta (G^2 + cN / m^2)}{2\lambda_{\min}(Q) - \beta\lambda^2_{\min}(Q)} \right) + \frac{\beta (G^2 + cN / m^2)}{2\lambda_{\min}(Q) - \beta\lambda^2_{\min}(Q)}.
\end{align}
where $\beta$ can be set as $\beta = \eta / N^{k}$, for $\forall ~k \in (0,1)$, since 
\begin{align}
\lim_{N \rightarrow +\infty}\left(1 - 1 / N^k\right)^{2N} = 1
\end{align}
for $k \in [1,+\infty)$.
\end{proof}

\subsection{Proof of Theorem \ref{thm:ut2}}
\label{subsec:pro_thmut2}

\begin{proof}
Under $\beta_i = \frac{\eta}{\lambda_{\min}(Q)i}$, \eqref{eq:finalineq} leads to
\begin{align}
& \mathbf{E}[\|\xi_{i + 1}\|_2^2] \\
\leq & \left(1 - \frac{\eta}{i}\right)^2 \mathbf{E}[\|\xi_i\|_2^2] + \frac{\eta^2(G^2 + cN / m^2)}{\lambda^2_{\min}(Q)i^2}
\label{eq:variableineq}
\end{align}
Let $H(\eta) = \max\left\{\|\xi_{1}\|_2^2, \frac{\eta^2 (G^2 + cN / m^2)}{(\eta - 1)\lambda^2_{\min}(Q)}\right\}$, so that $\mathbf{E}[\|\xi_{i}\|_2^2] \leq H(\eta) / i$
by induction. First, note that $\mathbf{E}[\|\xi_{1}\|_2^2] \leq H(\eta)$. So, if we assume that the convergence rate holds with $i$, we only need to show that it holds with $i + 1$. By \eqref{eq:variableineq}, we have
\begin{align}
& \mathbf{E}[\|\xi_{i + 1}\|_2^2] \\
\leq & \left(1 - \frac{\eta}{i}\right) \frac{H(\eta)}{i} + \frac{(\eta - 1)H(\eta)}{i^2} \\
\leq & \frac{(i - 1)H(\eta)}{t^2} \leq \frac{H(\eta)}{i + 1}.
\end{align}

Thus, we obtain the rate of convergence with diminishing stepsize as
\begin{align}
& \mathbf{E}[\|\xi_{N + 1}\|_2^2] \\
\leq &  \frac{\max\left\{\|\xi_{1}\|_2^2, \frac{\eta^2 (G^2 + cN / m^2)}{(\eta - 1)\lambda^2_{\min}(Q)}\right\}}{N} \\
\leq &  \frac{1}{N}\max\left\{\|\xi_{1}\|_2^2, \frac{\eta^2 G^2}{(\eta - 1)\lambda^2_{\min}(Q)}\right\} + \frac{\eta^2 c}{(\eta - 1)\lambda^2_{\min}(Q)}.
\end{align}
\end{proof}

\section{Extra Figures}
\label{sec:appfig}

In this section, we provide extra figures from out experiments. 





\begin{figure}[htb]
\centering
\begin{subfigure}[htb]{0.245\linewidth}
\centering
\includegraphics[width=\linewidth]{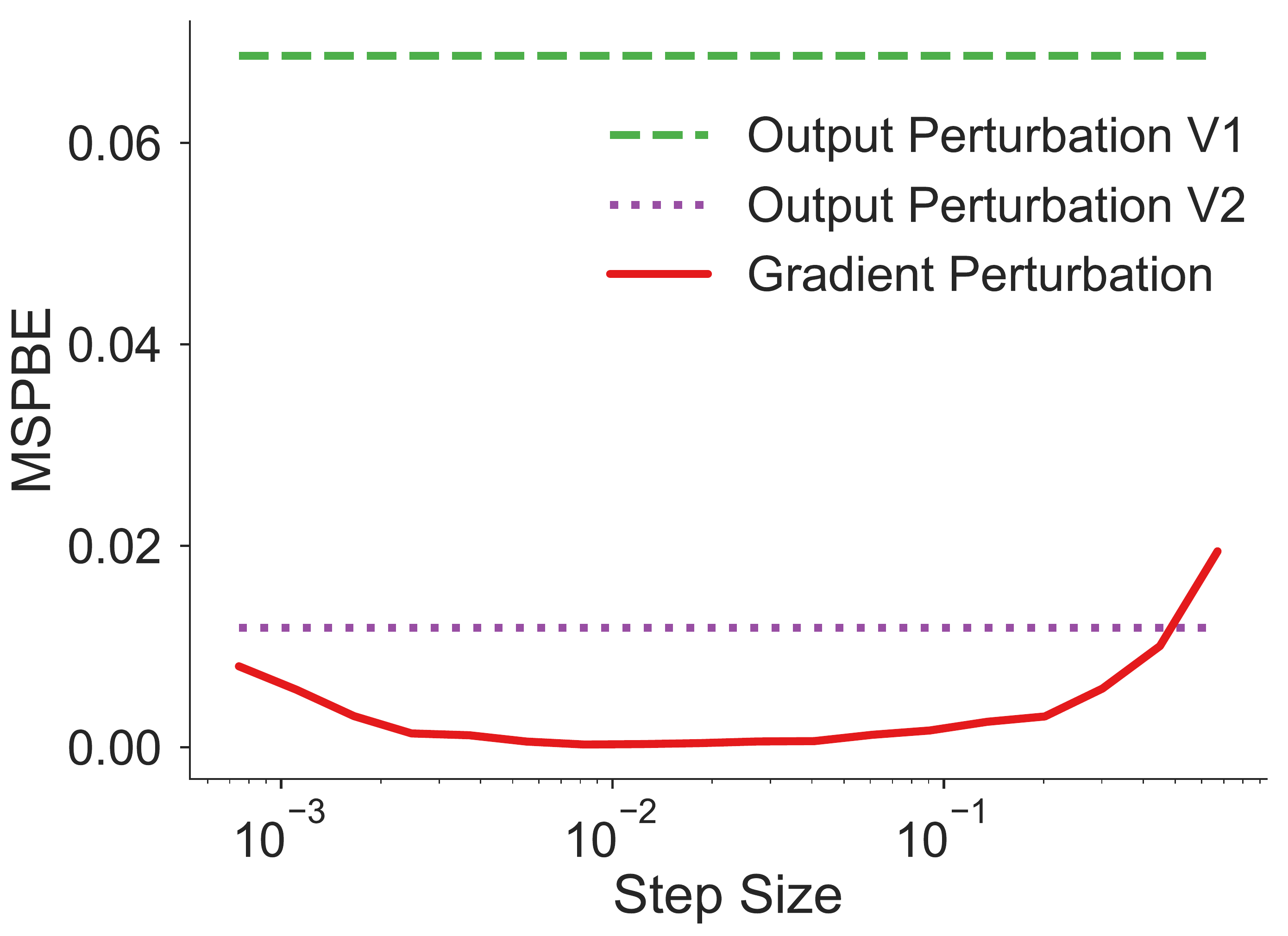}
\caption{$m = 1 \times 10^5$}
\label{fig:chain_varistep_10k}
\end{subfigure}
\begin{subfigure}[htb]{0.245\linewidth}
\centering
\includegraphics[width=\linewidth]{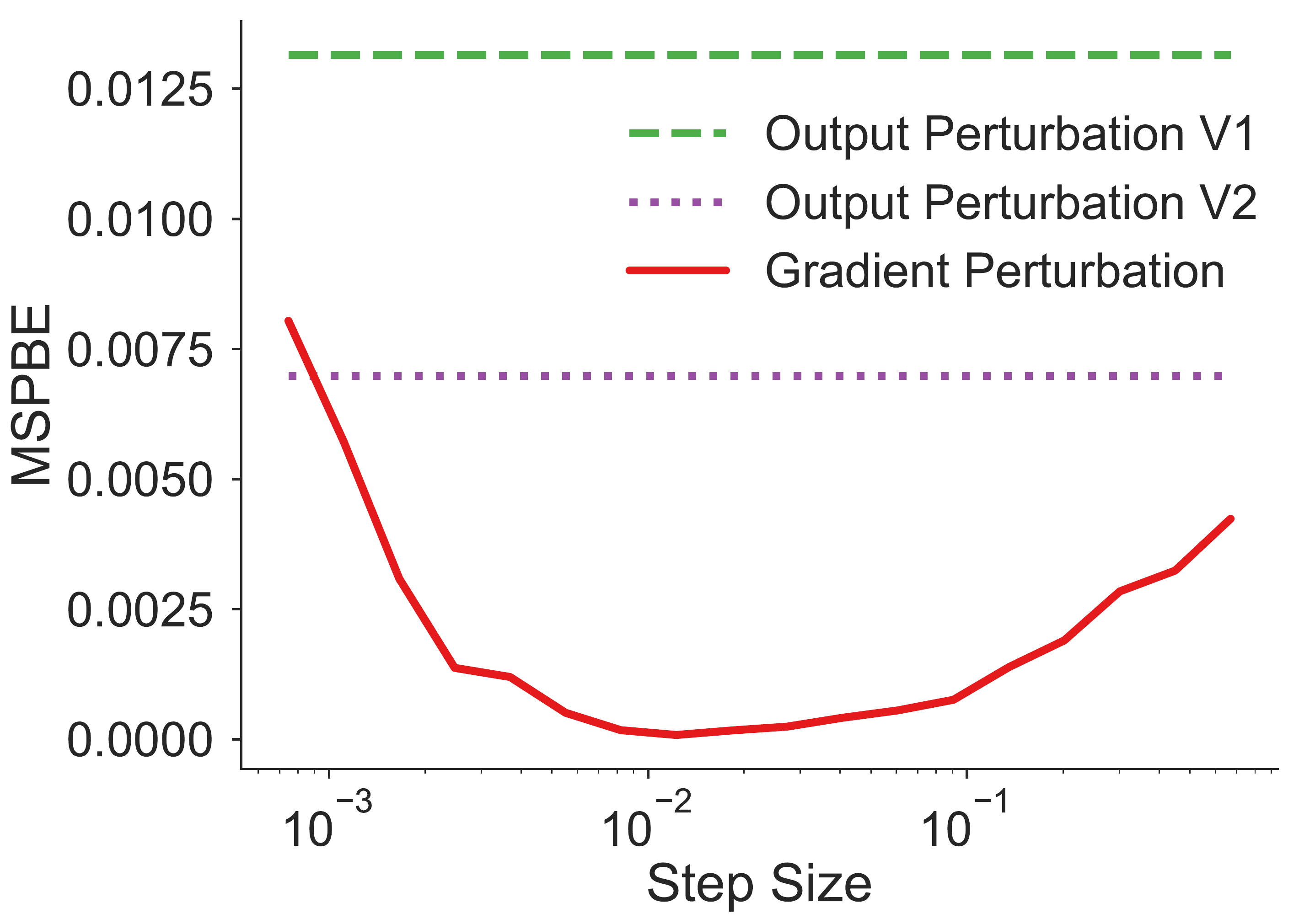}
\caption{$m = 2 \times 10^5$}
\label{fig:chain_varistep_20k}
\end{subfigure}
\begin{subfigure}[htb]{0.245\linewidth}
\centering
\includegraphics[width=\linewidth]{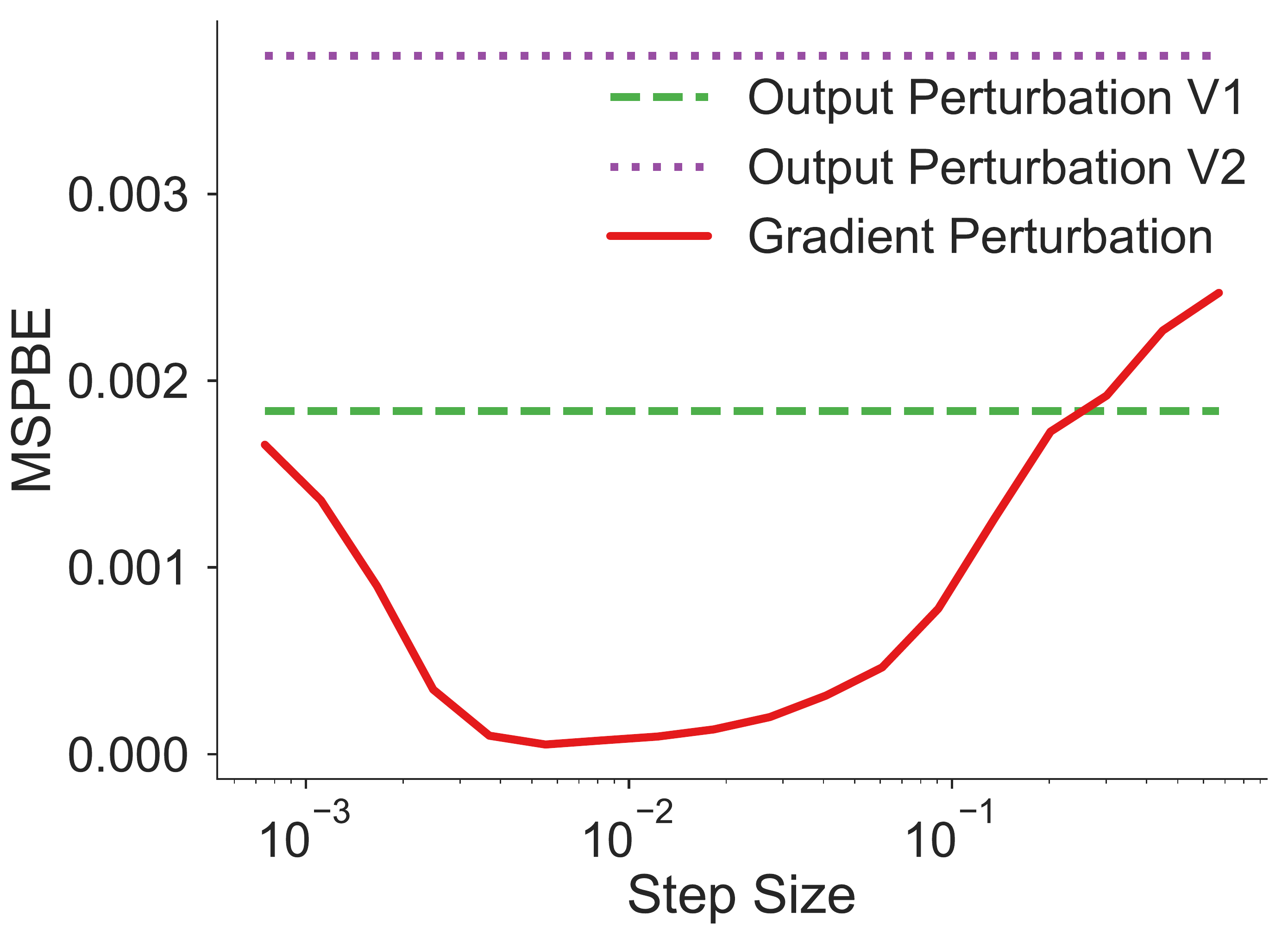}
\caption{fix $m = 4 \times 10^5$}
\label{fig:chain_varistep_40k}
\end{subfigure}
\begin{subfigure}[htb]{0.245\linewidth}
\centering
\includegraphics[width=\linewidth]{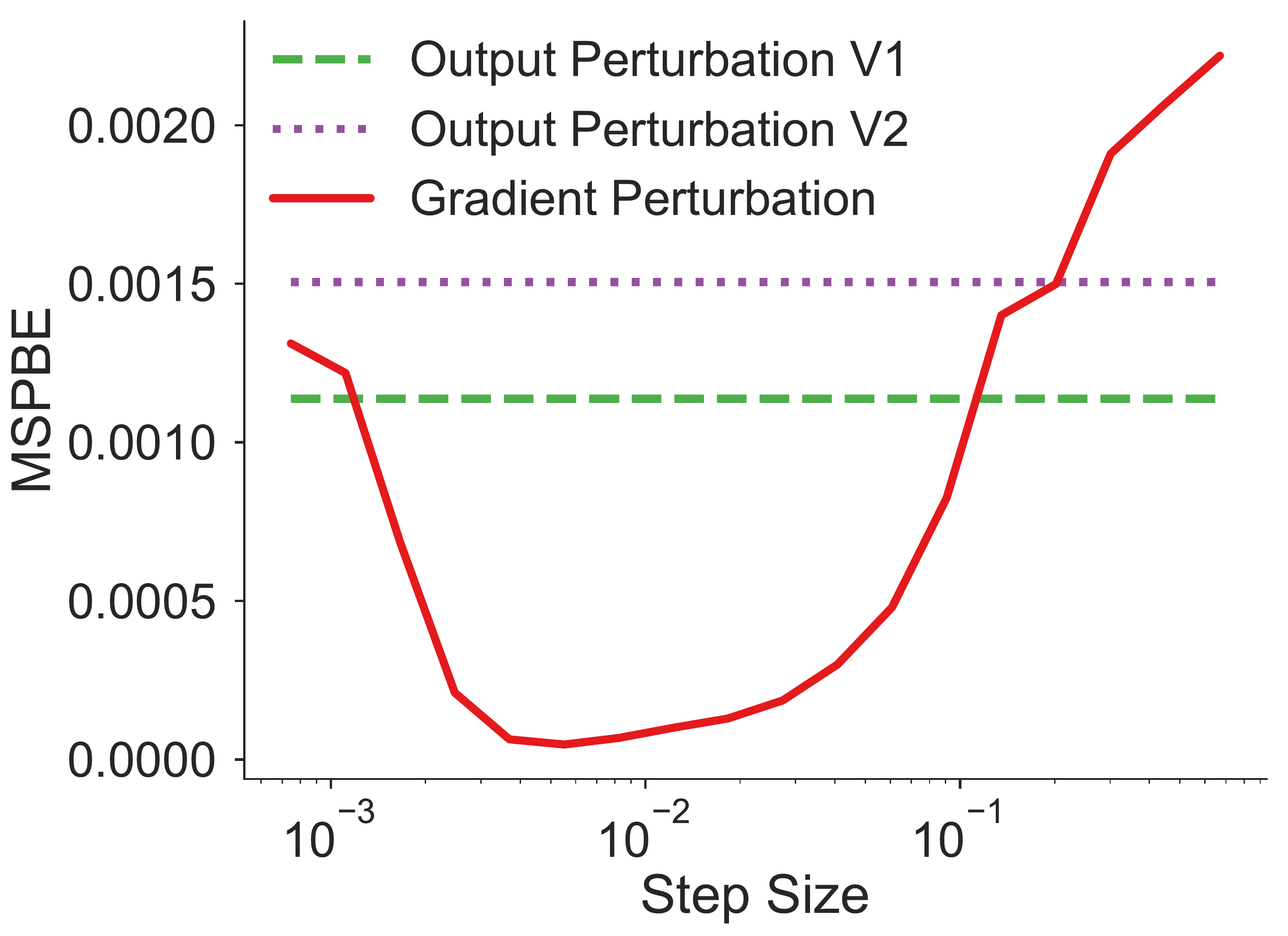}
\caption{$m = 5 \times 10^5$}
\label{fig:chain_varistep_50k}
\end{subfigure}
\caption{Sensitivity of step sizes in on-policy chain}
\label{fig:sensitivity-chain}
\end{figure}

Figure \ref{fig:sensitivity-chain} shows the sensitivity when the step size varies for the on-policy chain domain.

\begin{figure}[htb]
\centering
\begin{subfigure}[htb]{0.245\linewidth}
\centering
\includegraphics[width=\linewidth]{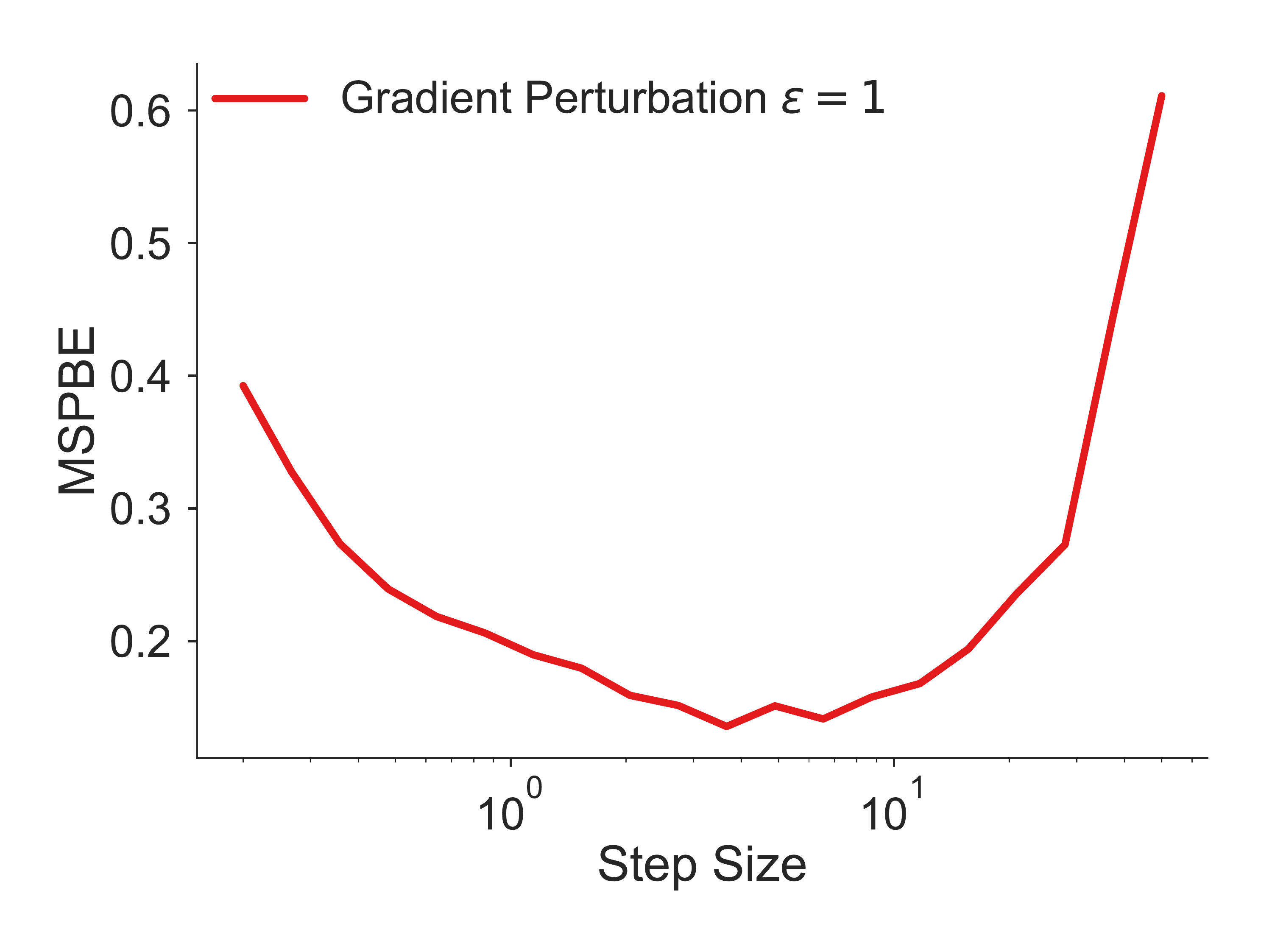}
\caption{$m = 1 \times 10^5$}
\label{fig:on_varistep_10k}
\end{subfigure}
\begin{subfigure}[htb]{0.245\linewidth}
\centering
\includegraphics[width=\linewidth]{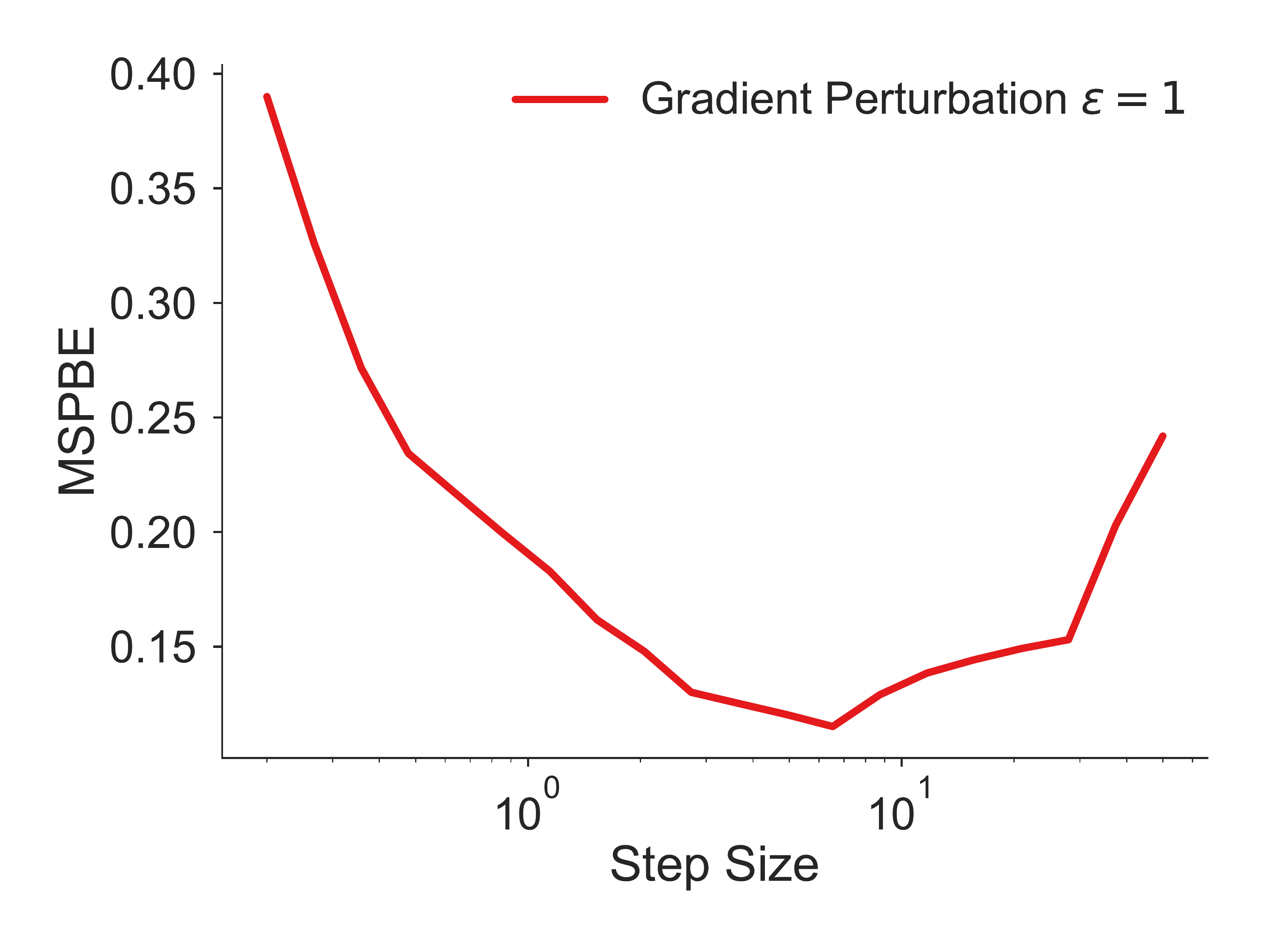}
\caption{$m = 2 \times 10^5$}
\label{fig:on_varistep_20k}
\end{subfigure}
\begin{subfigure}[htb]{0.245\linewidth}
\centering
\includegraphics[width=\linewidth]{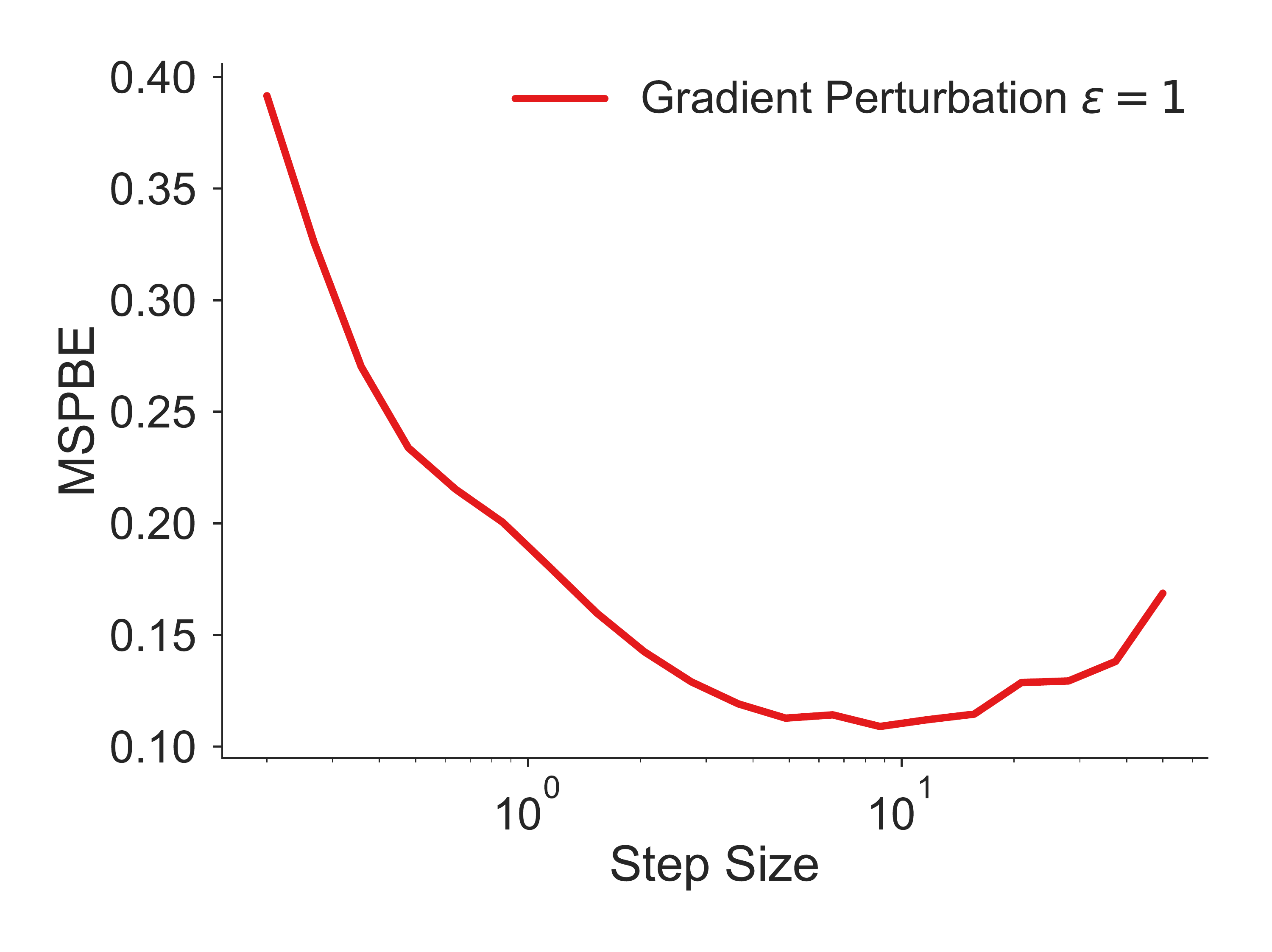}
\caption{fix $m = 4 \times 10^5$}
\label{fig:on_varistep_40k}
\end{subfigure}
\begin{subfigure}[htb]{0.245\linewidth}
\centering
\includegraphics[width=\linewidth]{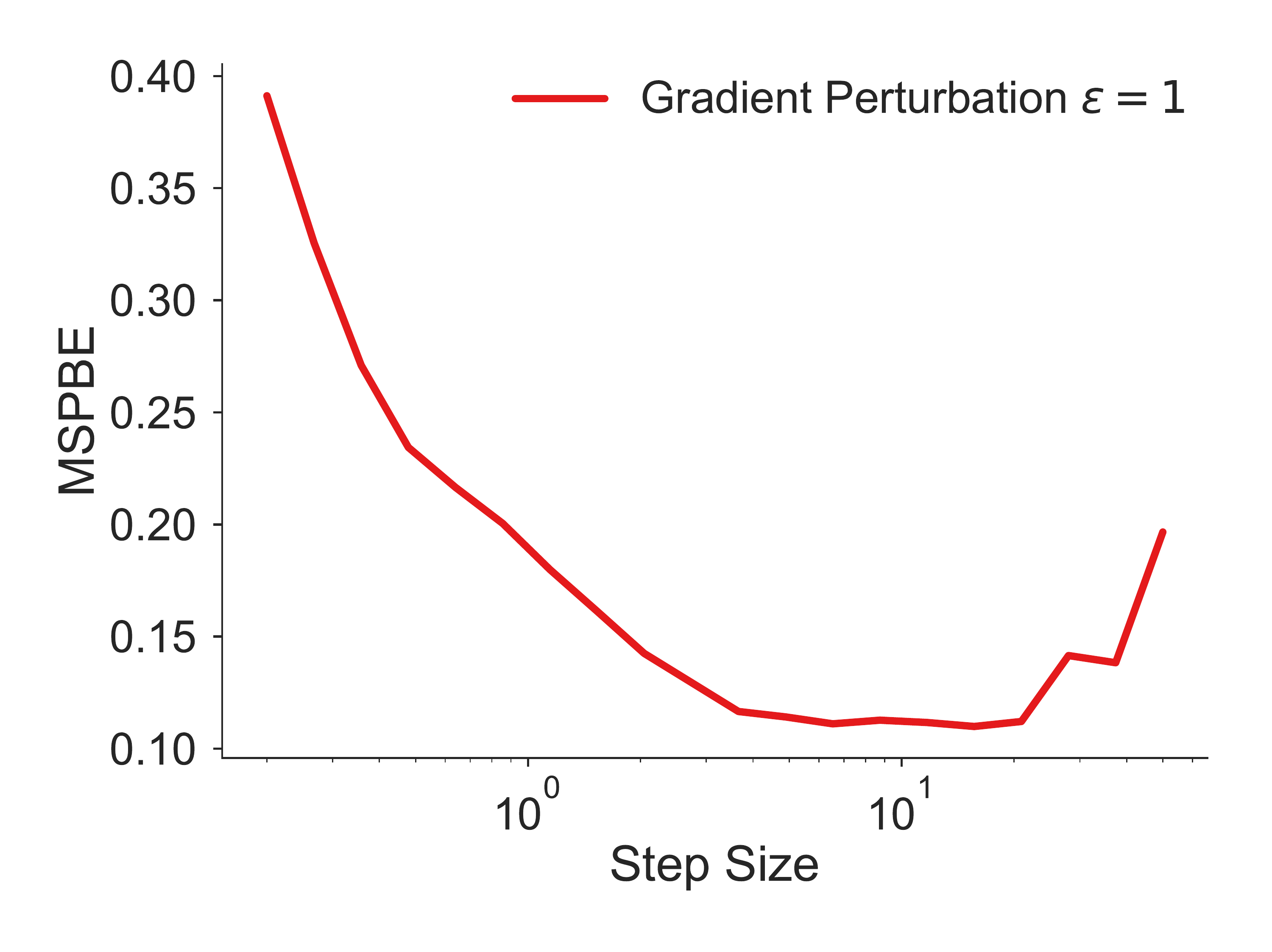}
\caption{$m = 5 \times 10^5$}
\label{fig:on_varistep_50k}
\end{subfigure}
\caption{Sensitivity of step sizes in on-policy mountain car}
\label{fig:sensitivity-car}
\end{figure}

Figure \ref{fig:sensitivity-car} shows the results of additional testing the sensitivity of our algorithm to the step size parameter.

\end{document}